\documentclass[10pt,twocolumn,letterpaper]{article}

\usepackage{cvpr}
\usepackage{fontawesome5}

\usepackage{microtype}

\makeatletter
\renewcommand{\paragraph}{\@startsection{paragraph}{4}{\z@}%
  {1.0ex \@minus .2ex}%
  {-1em}%
  {\normalfont\normalsize\bfseries}}
\renewcommand{\cvprsection}{\@startsection{section}{1}{\z@}%
  {-6pt \@minus 2pt}{5pt}{\large\bf}}
\renewcommand{\cvprsubsection}{\@startsection{subsection}{2}{\z@}%
  {-7pt \@minus 2pt}{4pt}{\elvbf}}
\renewcommand{\cvprsubsubsection}{\@startsection{subsubsection}{3}{\z@}%
  {-6pt \@minus 2pt}{3pt}{\tenbf}}
\makeatother
\usepackage{placeins}

\usepackage{colortbl}
\definecolor{acushead}{RGB}{232,238,244}
\definecolor{acusrow}{RGB}{239,247,248}

\definecolor{cvprblue}{rgb}{0.21,0.49,0.74}
\usepackage[pagebackref,breaklinks,colorlinks,allcolors=cvprblue]{hyperref}

\def\confName{CVPR}
\def\confYear{2027}

\makeatletter
\def\@maketitle{%
  \newpage
  \null
  \vskip 0.12in
  \begin{center}%
    {\LARGE\bfseries \@title \par}%
    \vskip 16pt
    {\large \@author \par}%
  \end{center}%
  \vskip 6pt
}
\makeatother

\title{ACUS: An Angular Calibrated Uncertainty Score for Incomplete-View CT}

\author{%
Zihan Deng$^{1}$ \quad Chuanzhi Xu$^{2}$ \quad Quankeng Huang$^{1}$ \quad Yujie Liu$^{3}$\\[0.45em]
Zhenzhen Gong$^{3}$ \quad Junjie Huang$^{4}$ \quad Tianye Niu$^{4,5}$ \quad Lequan Yu$^{1}$\\[0.7em]
{\normalsize $^{1}$The University of Hong Kong \quad $^{2}$The University of Sydney \quad $^{3}$Chongqing University}\\[0.28em]
{\normalsize $^{4}$University of Science and Technology of China \quad $^{5}$Shenzhen Bay Laboratory}\\[0.5em]
\href{https://github.com/FrankDengAI/ACUS}{\faGithub}%
}

\begin{document}
\raggedbottom
\setlength{\parskip}{0pt}
\maketitle

\begin{abstract}
Sparse-view and limited-arc computed tomography (CT) are commonly treated as projection completion: a network fills an unmeasured view, and the result is ranked by similarity to one ground-truth image.
That score can be high even when the query angle is far from every acquired measurement, because a learned prior may invent anatomy the scan never constrained.
Existing protocols therefore leave a basic question unanswered: how does predictive reliability change with distance from the acquired views?
We propose the Angular Calibrated Uncertainty Score (ACUS) to answer that question.
ACUS is the Winkler interval score of a nominal $90\%$ predictive interval on held-out projections, binned by distance to the nearest acquired view.
It charges interval width and adds a large miss penalty when a held-out pixel falls outside the interval, so a narrow band helps only when coverage holds.
We use collapse on acquired angles as a necessary consistency check on the predictor.
On chest and dental cone-beam orbits, interval width grows with the angular gap. Empirical coverage ranges from $77\%$ to $92\%$, and the peak signal-to-noise ratio (PSNR) of the point estimate falls as the width rises.
Nearest-neighbor, interpolation, and point completion, each given one residual scale, have zero reported far coverage and incur outside-distance penalties.
ACUS exposes gap-dependent overconfidence that a point-estimate PSNR alone does not measure.
\end{abstract}

\section{Introduction}
\label{sec:intro}

\begin{figure}[t]
\centering
\includegraphics[width=\columnwidth]{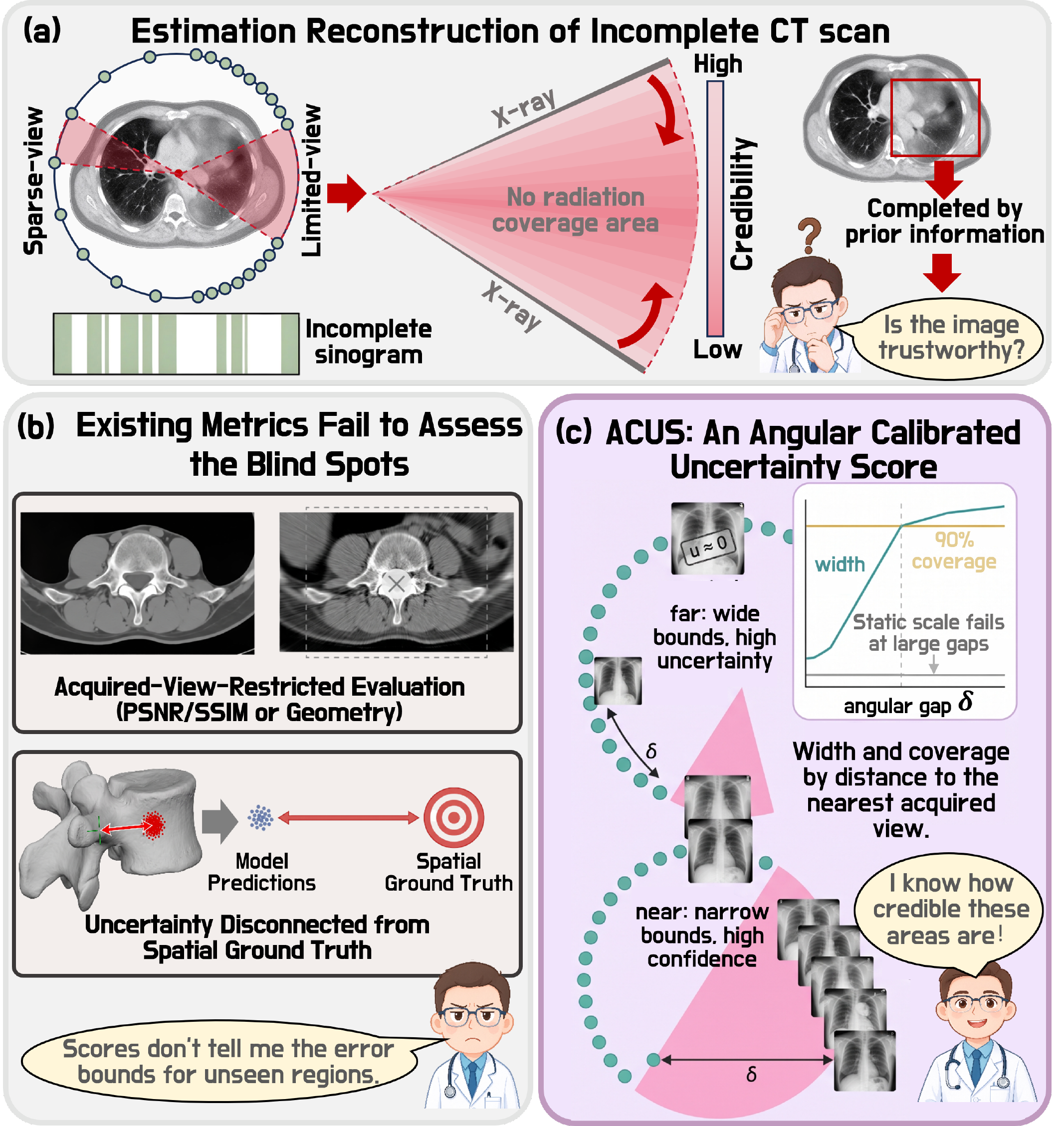}
\caption{Incomplete-view CT as an interval problem.
(a) Sparse-view and limited-arc scans leave angles with no ray coverage.
A prior can complete them, and the question is whether the image is trustworthy.
(b) PSNR, SSIM, and acquired-view geometry give no error bound on those angles.
(c) ACUS reports width and coverage by the gap to the nearest acquired view: narrow nearby, wide far away.
A static scale fails at large gaps.}
\label{fig:idea}
\end{figure}

Incomplete-view computed tomography (CT) includes sparse-view and limited-arc acquisition.
\Cref{fig:idea}(a) shows the incomplete sinogram: angles with no rays can still be filled from an anatomical prior~\cite{jin2017fbpconvnet,yang2026xwin}.
The result can look anatomical, and the question is whether it is trustworthy where the scan never measured.

\Cref{fig:idea}(b) shows why the peak signal-to-noise ratio (PSNR) or structural similarity (SSIM) does not answer that question.
Either score compares a prediction with one reference and can stay high far from every acquired view, because the prior can introduce structure the measurements do not determine.
A geometry summary of the acquired trajectory has the same blind spot: it describes the rays that were taken and gives no error bound on the held-out projection~\cite{kak2001,natterer2001}.
Many volumes match the acquired integrals and disagree elsewhere.
This set is the measurement fiber.

\Cref{fig:idea}(c) scores that held-out projection.
ACUS is the Winkler interval score of a nominal $90\%$ interval, indexed by the gap $\delta$ to the nearest acquired view~\cite{winkler1972,gneiting2007}.
The interval is narrow near an acquired view and wide far away, and coverage is reported with the width.
A static scale, one residual magnitude at every angle, fails at large gaps.
Collapse on acquired angles checks that compatible volumes reproduce the measurements that were taken.
The score charges width and penalizes pixels outside the interval, so a narrow interval helps only when coverage holds.
\Cref{sec:fiber} defines the score, and \Cref{sec:volume} records a refinement that can raise volume similarity while agreeing less with the acquired projections.

In this paper, we propose ACUS, a gap-indexed interval score for incomplete-view CT, together with a residual-scale estimator and a held-out evaluation on chest and dental cone-beam orbits.
Our contributions can be summarized as:
\begin{itemize}
\item \textbf{Angular interval score.}
ACUS is the nominal $90\%$ Winkler score on held-out projections, binned by the angular gap $\delta$.
The score charges interval width and a miss penalty, so a narrow interval is rewarded only when coverage holds.
Acquired-view collapse is the consistency check.
\item \textbf{Residual-scale estimator.}
We train a feed-forward mean and pixelwise scale with held-out likelihood and an acquired-view collapse gate.
The published interval opens with the angular gap, instead of reusing the static scale in \Cref{fig:idea}(c).
\item \textbf{Cone-beam evaluation.}
We evaluate the score and the estimator on chest and dental cone-beam orbits under sparse and multi-arc $K{=}8$ sampling, with 40 test cases per domain.
Interval width grows with the gap, empirical coverage ranges from $77\%$ to $92\%$, and constant-scale point predictors fail the far bins by missing held-out observations.
\end{itemize}

\section{Related work}
\label{sec:related}

\paragraph{Reconstruction and projection prediction.}
Incomplete-view tomography admits multiple volumes consistent with the acquired line integrals~\cite{kak2001,natterer2001}.
Learned reconstruction can select a useful estimate through an anatomical prior. Convolutional post-processing~\cite{jin2017fbpconvnet} and dual-domain metal-artifact reduction~\cite{lin2019dudonet} illustrate how image and projection information can improve point estimates, although their acquisition and corruption models differ from ours.
Implicit attenuation fields~\cite{zha2022naf,cai2024saxnerf} and radiative Gaussian representations~\cite{cai2024xgaussian,zha2024r2gaussian} instead fit a spatial representation through rendered projections.
These advances motivate a complementary evaluation question: how reliable is a predicted line integral at an angle absent from the scan?
A strong reconstruction score does not by itself answer that question.

\paragraph{Predictive representations.}
Masked image modeling~\cite{he2022mae} and joint-embedding prediction~\cite{assran2023ijepa} learn transferable features rather than calibrated detector intensities.
X-WIN applies viewpoint-conditioned latent prediction to chest radiographs and demonstrates projection rendering~\cite{yang2026xwin}.
This makes predictive representations relevant backbones, but the latent objective does not specify a nominal interval or a penalty for missing a held-out measurement.
Our point-completion baseline isolates a point-estimate objective using the same backbone and an $\ell_1$ projection loss. It is not a reproduction of every cited representation learner.

\paragraph{Uncertainty and its evaluation target.}
Geometric incompleteness quantifies the information supplied by a source trajectory at a voxel~\cite{clackdoyle2020}.
Uncertainty-aware null-space networks combine data-consistent reconstruction with an image-domain residual scale~\cite{angermann2023}.
Recent analysis of radiative Gaussian uncertainty distinguishes posterior spread from localization of realized reconstruction error~\cite{zhao2026variance}.
The target matters in each case: geometry, voxel error, and a held-out projection are different quantities.
ACUS evaluates the last, retaining both interval length and outside distance through the classical interval score~\cite{winkler1972,gneiting2007}.
Distribution-free image regression uses a separate calibration procedure to obtain coverage guarantees under its assumptions~\cite{angelopoulos2022imaging}.
Our untempered Gaussian intervals receive an empirical assessment, not such a guarantee. A scoring rule determines what predictions are rewarded, while calibration determines how interval frequencies behave on a specified population. Keeping these roles separate allows the same interval criterion to evaluate different predictors without attributing statistical guarantees to their architecture.

\paragraph{Sequential acquisition.}
Learned sampling policies~\cite{wang2023activect}, sequential experimental design~\cite{wang2024tci}, and stopping policies~\cite{wang2025stop} optimize the acquisition sequence or its termination.
Perturbed Gaussian Ensemble uses structural variation across density perturbations to select a new view~\cite{wu2026pge}.
Those methods evaluate the utility of acquiring a measurement. ACUS instead scores the predictive interval before that measurement is revealed.
The distinction permits retrospective comparison without assuming that an uncertainty ranking is already a validated acquisition policy.
Testing whether the score improves a closed-loop policy remains a separate experiment.

\section{Method}
\label{sec:acus-method}

\begin{figure*}[t]
\centering
\includegraphics[width=\textwidth]{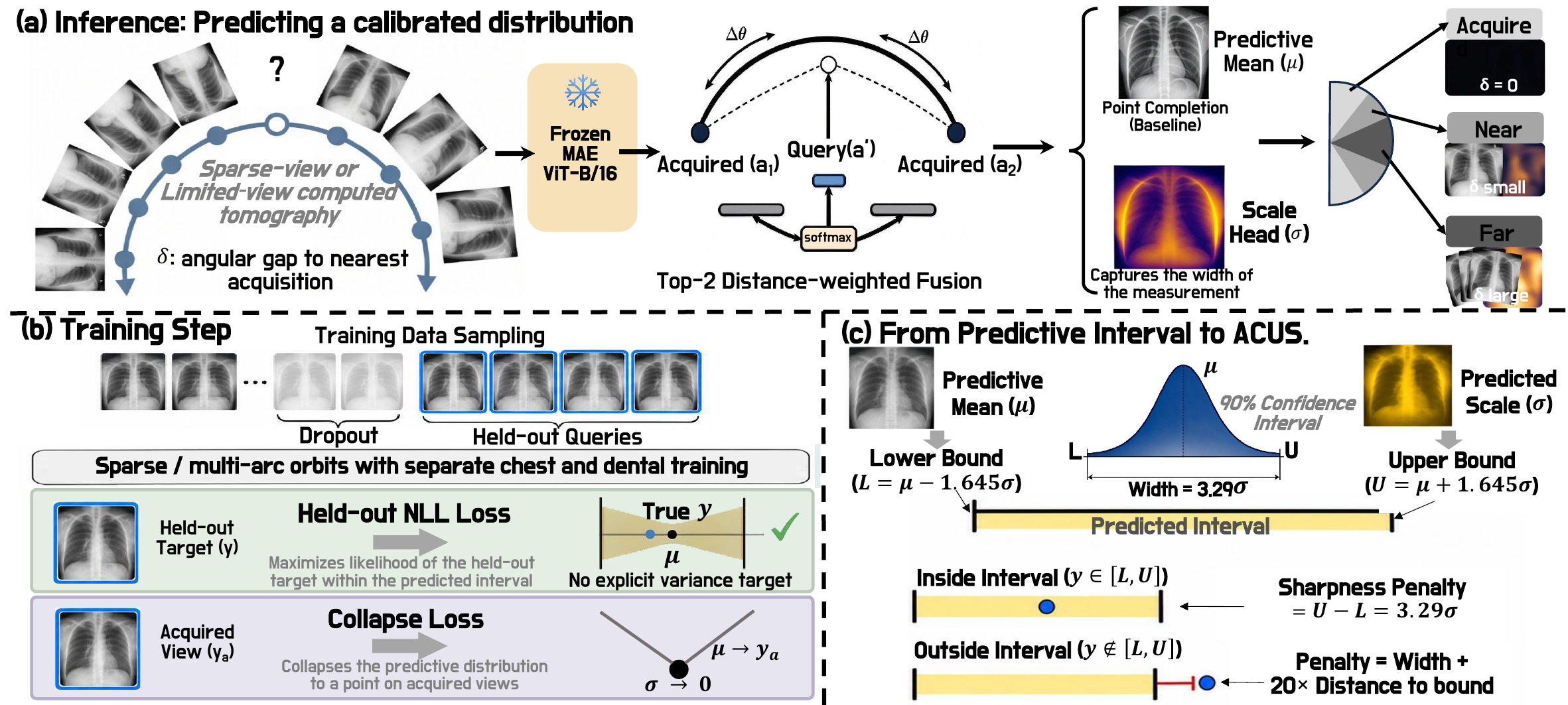}
\caption{Interval estimator used to compute ACUS.
(a) A frozen MAE encodes acquired views.
Distance-weighted softmax fuses the two nearest into a mean $\mu$ and a scale $\sigma$.
Point completion is the mean-only baseline.
$\delta$ marks acquired, near, and far queries.
(b) Chest and dental specialists train on sparse and multi-arc orbits with dropout and held-out queries.
Gaussian likelihood has no variance target.
Collapse drives $\sigma$ toward zero and $\mu$ toward the acquired view.
(c) The nominal $90\%$ interval is $\mu\pm 1.645\sigma$ (width $3.29\sigma$).
Inside, ACUS equals the width.
Outside, it adds twenty times the distance to the bound.}
\label{fig:method}
\end{figure*}

ACUS scores the nominal $90\%$ interval of a predictive mean and a pixelwise scale.
On a covered pixel the score equals the interval width.
On a missed pixel it adds twenty times the outside distance, so a narrow interval helps only when coverage holds (\Cref{eq:decompose}).
Held-out likelihood trains the scale to open with the angular gap.
A collapse loss trains it to close on acquired angles.
\Cref{fig:method}(a) is the predictor, (b) its training, and (c) the score.
The gap and the score are defined below.
The network reads only the two nearest views.
$\delta$ and the measurement fiber still depend on the full acquired set.

\subsection{Angular interval scoring}
\label{sec:fiber}

A circular cone-beam orbit provides projections $y_a = P_a(V)$ of an unknown volume $V$.
One scan reveals a set $\mathcal{A}$.
In the main protocol $\mathcal{A}$ is either a sparse subset of $K{=}8$ views on a 50-view, $180^\circ$ orbit, or two to three short arcs of $25^\circ$ to $45^\circ$ whose total view count is still $K{=}8$.
A single contiguous arc of $50^\circ$ to $120^\circ$ is reported only as a stress test.
The measurement fiber is
\begin{equation}
\mathcal{F}(\mathcal{A}, y_{\mathcal{A}})
  = \{ V : P_a(V) = y_a,\; a\in\mathcal{A} \}.
\label{eq:fiber}
\end{equation}
At a query $a'\notin\mathcal{A}$ the projections still compatible with the acquired integrals are the push-forward $\{P_{a'}(V): V\in\mathcal{F}\}$.
The fiber is a set of volumes.
It carries no probability by itself, and the network never enumerates it.
On an acquired angle the noise-free push-forward is the singleton $\{y_a\}$.

\paragraph{What the angular gap controls.}
For one fixed volume the cone-beam transform is continuous in the source angle, so the true projection returns to the acquired measurement as the query approaches it.
That fact does not bound the rest of $\mathcal{F}$.
Finitely many views leave a null space, and without a regularity class two consistent volumes can separate after an arbitrarily small angle step.
Inside a class $\mathcal{B}$ on which the transform is uniformly continuous in angle, every legal projection at $a'$ lies near the nearest acquired measurement, and the diameter of this restricted push-forward tends to zero with the gap $\delta(a')$.
More precisely, if $\|P_b(V)-P_a(V)\|\leq\omega_{\mathcal B}(|b-a|)$ uniformly over $V\in\mathcal B$, the triangle inequality gives diameter at most $2\omega_{\mathcal B}(\delta)$ for the restricted push-forward.
This conditional observation motivates the boundary check, not a numerical bound on the network's scale.
It does not set how fast the spread opens at a large gap. That rate depends on $\mathcal{B}$.
In practice the specialist learns an anatomical prior from training data. We do not identify its support with a known regularity class or estimate a uniform modulus from finite tests.
We use $\delta(a')=\min_{a\in\mathcal A}|a'-a|$ on the sampled $180^\circ$ trajectory, without identifying its endpoints.
\Cref{fig:method}(a) marks three regimes of this gap: acquired ($\delta{=}0$), near (small $\delta$), and far (large $\delta$).
The numerical cuts, under $20^\circ$ and at least $40^\circ$, are fixed in \Cref{sec:exp}.
The index is computed from the mask, so it is independent of the predictor's errors.
Pooling solely by view count would mix interpolation near measured support with extrapolation across large gaps.
It is not a voxel-incompleteness score and not a condition number of the whole missing wedge, so sparse and multi-arc orbits are reported separately at the same $\delta$.

\paragraph{The published interval.}
\Cref{fig:method}(c) draws the interval that ACUS scores.
We publish a diagonal Gaussian on detector pixels, as a predictive stand-in for the restricted push-forward,
\begin{equation}
p(x\mid a',\mathcal{A},y_{\mathcal{A}})
  = \mathcal{N}\!\big(\mu(a'),\,\mathrm{diag}\,\sigma(a')^{2}\big),
\label{eq:gauss}
\end{equation}
where $x$ is the predicted detector image and $\Omega$ is the set of detector pixels. We summarize the distribution by the mean scale
\begin{equation}
u(a') = \tfrac{1}{|\Omega|}\sum_{i\in\Omega}\sigma_i(a').
\label{eq:width}
\end{equation}
The conditioning in \eqref{eq:gauss} is written on all of $\mathcal{A}$.
\Cref{sec:method} uses only the two nearest views.
The nominal $90\%$ interval at detector pixel $i$ is $[L_i,U_i]=\mu_i\pm 1.645\,\sigma_i$.
ACUS is the mean Winkler score of that interval on the held-out intensity $y_i$~\cite{winkler1972,gneiting2007},
\begin{equation}
S_i =
\begin{cases}
U_i-L_i, & L_i\le y_i\le U_i,\\[1pt]
U_i-L_i+\dfrac{2}{\alpha}(L_i-y_i), & y_i<L_i,\\[1pt]
U_i-L_i+\dfrac{2}{\alpha}(y_i-U_i), & y_i>U_i,
\end{cases}
\label{eq:winkler}
\end{equation}
\begin{equation}
\mathrm{ACUS}(a')=\tfrac{1}{|\Omega|}\sum_{i\in\Omega} S_i,
\qquad \alpha=0.1.
\label{eq:acus}
\end{equation}
Lower is better.
With $\alpha=0.1$ the central interval is $\mu\pm 1.645\,\sigma$, so the width $U_i-L_i=3.29\,\sigma_i$ is the sharpness penalty in \Cref{fig:method}(c).
A miss adds $20$ times its distance outside that bound.
For a continuous target cumulative distribution $F$, differentiating the expected score with respect to its two endpoints gives
\begin{equation}
 \partial_L\mathbb E S=-1+20F(L),\quad
 \partial_U\mathbb E S=1-20[1-F(U)].
\label{eq:proper}
\end{equation}
The optima are the $0.05$ and $0.95$ quantiles. They are unique when these quantiles are unique~\cite{winkler1972,gneiting2007}.
Thus the score elicits the central interval, rather than the full predictive density.
The Gaussian restriction can prevent $\mu\pm1.645\sigma$ from attaining those quantiles for an asymmetric target.
Properness is an expected-score statement under the target distribution, not a claim that a fitted interval is calibrated.
It does not identify $\mathcal{F}$.
Let $d_i=(L_i-y_i)_++(y_i-U_i)_+$ and let $\bar d$ be its mean.
The score then splits as
\begin{equation}
\mathrm{ACUS}(a')
  = 3.29\, u(a') + 20\,\bar d(a').
\label{eq:decompose}
\end{equation}
On a covered pixel the score equals the interval length $3.29\,\sigma_i$, not the scale $\sigma_i$.
If every pixel is covered, $\mathrm{ACUS}=3.29\,u$.
Tables list the mean scale $\bar u(\delta)$ and the coverage $\bar c(\delta)$ so the two terms stay visible.
Coverage and mean scale must be read together. Coverage near $90\%$ alone does not bound the miss term, because misses can have different magnitudes.
PSNR of $\mu$ is written beside them and is not the acceptance test.

A held-out projection is one legal member of the push-forward, the projection of the scanned volume.
Coverage is the rate at which that member falls inside the interval.
A rate near $90\%$ is evidence of marginal empirical coverage on the evaluated pixels. It does not establish conditional coverage for every case, angle, or anatomical structure.
It does not measure the diameter of $\{P_{a'}(V):V\in\mathcal{F}\}$.
The acquired-angle gate checks the boundary where the restricted push-forward is a point. Passing this check does not certify that an interval contains every compatible projection.
An undersized scale incurs the miss term whenever a held-out intensity falls outside the interval, regardless of its point-estimate PSNR.
Small $u$ with coverage holding means $\mu$ is a usable point readout.
Large $u$ indicates predictive spread within the learned anatomy class, which a mean image alone omits.
Small $u$ with failed coverage indicates an overconfident predictor, as in the dental-to-chest case.
Pixelwise independence is a modeling choice, and a draw from \eqref{eq:gauss} is not a projection of a volume in $\mathcal{F}$.
A volume readout $V^\star$ would also have to match $y_a$ on $\mathcal{A}$ and land in a high-probability region of $p$ at the query.
\Cref{sec:volume} records a volume readout that leaves the fiber.

\subsection{Residual-scale learning and collapse}
\label{sec:method}

\begin{table}[t]
\centering
\caption{Sparse $K{=}8$, 40 test cases per domain. Entries are near/far. $u$ is mean scale, not interval length. Cover targets $0.90$. Point denotes point completion. ACUS scale standard deviations (near/far): Chest $0.003/0.003$, Dental $0.002/0.002$.}
\label{tab:sparse}
\scriptsize
\setlength{\tabcolsep}{8.50301pt}
\begin{tabular*}{\columnwidth}{lccc}
\toprule
\rowcolor{acushead}Method & $u$ near/far & Cover near/far & PSNR near/far \\
\midrule
\rowcolor{acushead}\multicolumn{4}{l}{\textbf{Chest}} \\
Nearest & $0.014/0.014$ & $0.61 / 0.00$ & $30.27 / 16.40$ \\
Linear & $0.012/0.012$ & $0.70 / 0.00$ & $32.34 / 16.34$ \\
Point ($\ell_1$) & $0.013/0.013$ & $0.72 / 0.00$ & $32.17 / 14.14$ \\
\midrule
\rowcolor{acusrow}
ACUS & $0.018/0.088$ & $0.81 / 0.87$ & $\mathbf{32.48} / \mathbf{18.64}$ \\
\midrule
\rowcolor{acushead}\multicolumn{4}{l}{\textbf{Dental}} \\
Nearest & $0.014/0.014$ & $0.67 / 0.00$ & $30.24 / 18.51$ \\
Linear & $0.012/0.012$ & $0.76 / 0.00$ & $31.72 / 18.54$ \\
Point ($\ell_1$) & $0.011/0.011$ & $0.73 / 0.00$ & $31.78 / 20.95$ \\
\midrule
\rowcolor{acusrow}
ACUS & $0.024/0.061$ & $0.92 / 0.86$ & $\mathbf{31.90} / \mathbf{21.12}$ \\
\bottomrule
\end{tabular*}
\end{table}

\begin{table}[t]
\centering
\caption{Multi-arc $K{=}8$, 40 test cases per domain. Entries are near/far. $u$ is mean scale, not interval length. Cover targets $0.90$. Point denotes point completion. ACUS scale standard deviations (near/far): Chest $0.003/0.008$, Dental $0.001/0.002$. Acquired scale is $0.002$ in both domains (PSNR $52.6/54.9$ dB).}
\label{tab:arc}
\scriptsize
\setlength{\tabcolsep}{8.50301pt}
\begin{tabular*}{\columnwidth}{lccc}
\toprule
\rowcolor{acushead}Method & $u$ near/far & Cover near/far & PSNR near/far \\
\midrule
\rowcolor{acushead}\multicolumn{4}{l}{\textbf{Chest}} \\
Nearest & $0.022/0.022$ & $0.74 / 0.00$ & $29.05 / 15.51$ \\
Linear & $0.022/0.022$ & $0.71 / 0.00$ & $29.18 / 15.28$ \\
Point ($\ell_1$) & $0.026/0.026$ & $0.77 / 0.00$ & $28.79 / 13.59$ \\
\midrule
\rowcolor{acusrow}
ACUS & $0.023/0.095$ & $0.77 / 0.80$ & $\mathbf{29.59} / \mathbf{17.48}$ \\
\midrule
\rowcolor{acushead}\multicolumn{4}{l}{\textbf{Dental}} \\
Nearest & $0.020/0.020$ & $0.79 / 0.00$ & $29.20 / 19.83$ \\
Linear & $0.021/0.021$ & $0.77 / 0.00$ & $29.32 / 19.80$ \\
Point ($\ell_1$) & $0.018/0.018$ & $0.70 / 0.00$ & $29.55 / \mathbf{22.47}$ \\
\midrule
\rowcolor{acusrow}
ACUS & $0.027/0.063$ & $0.89 / 0.91$ & $\mathbf{29.78} / 22.37$ \\
\bottomrule
\end{tabular*}
\end{table}

\paragraph{Backbone.}
\Cref{fig:method}(a) is this inference step.
Each acquired projection is a $224\times224$ cone-beam image.
A frozen MAE Vision Transformer (ViT-B/16) encoder~\cite{he2022mae} tokenizes it.
An action-conditioned transformer fuses the two nearest acquired views by distance-weighted softmax at temperature $0.10$, conditioned on the signed angle from the query.
One head reads the fused tokens as the predictive mean $\mu(a')$.
Point completion is that mean path trained with an $\ell_1$ projection loss and no scale head.
A second head reads the same tokens as a per-pixel log-scale, initialized near $\log 0.08$ and clamped so that $\sigma\in[e^{-6},0.35]$.
This scale head is the only module added to the point-completion backbone.
It is what allows the published width to vary with $\delta$.
The clamp stops the scale from saturating, and the log term in the likelihood penalizes inflation.
At test time $K$ is fixed at $8$ and every unobserved angle on the $50$-view orbit is queried.
$\delta$ is computed from every view in $\mathcal{A}$.
The other views constrain $\mathcal{F}$ and do not enter the features.
Near-angle mixing is off, so a small gap does not replace $\mu$ by a copy of the nearest view.

\begin{figure*}[t]
\centering
\includegraphics[width=\textwidth]{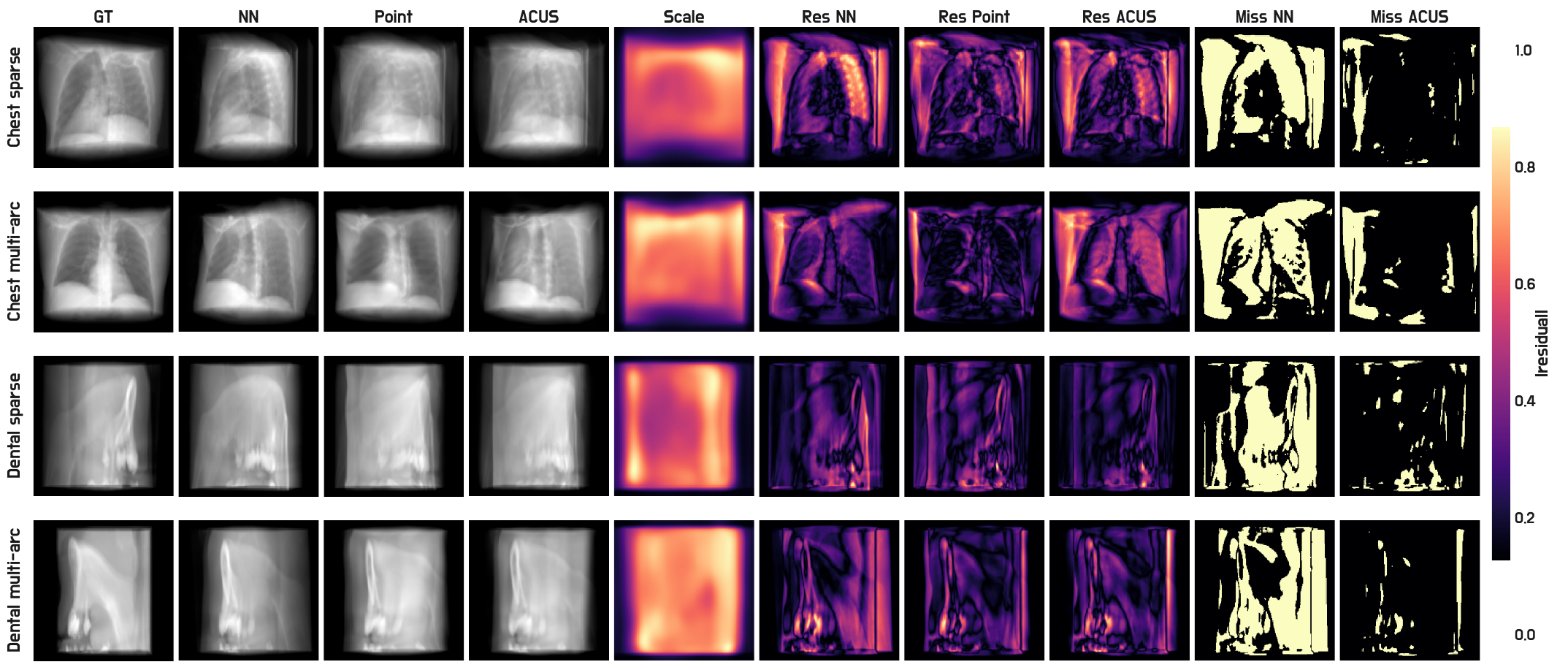}
\caption{Far-angle held-out projections on chest and dental cone-beam orbits.
Columns: ground truth (GT), nearest neighbor (NN), point completion, ACUS mean, ACUS scale, color residual maps for NN, point completion, and ACUS, and miss masks, with a shared residual colorbar.
Constant-scale baselines miss most far pixels. ACUS keeps a structured scale and a sparse miss mask.}
\label{fig:medfar}
\end{figure*}

\paragraph{Held-out likelihood.}
\Cref{fig:method}(b) draws $K$ from $6$ to $10$, takes a sparse subset with probability $0.55$ and a multi-arc subset otherwise, holds out four query views, and drops $15\%$ of the acquired views.
Chest and dental are separate specialists.
The held-out target is scored with the Gaussian negative log-likelihood (NLL)
\begin{equation}
\mathcal{L}_{\mathrm{NLL}}
  = \tfrac12\,\mathbb{E}\big[\log\sigma^{2} + (y-\mu)^{2}/\sigma^{2}\big].
\label{eq:nll}
\end{equation}
A light $\ell_1$ and gradient penalty on $\mu$ ($\lambda{=}0.15$ and $0.05$) keeps the mean from drifting while the scale is learning.
As marked in \Cref{fig:method}(b), the scale has no explicit variance target.
For a fixed conditioning input and mean, minimizing the expected NLL over an unconstrained scale yields $\sigma_i^2=\mathbb E[(Y_i-\mu_i)^2]$.
Optimizing the mean as well recovers the conditional mean and variance when finite and representable.
The scale therefore responds to both residual variability and mean bias. It is not a separately identified epistemic variance.
NLL trains the density parameters, whereas ACUS evaluates its interval endpoints. The auxiliary penalties and scale clamp further separate the fitted solution from this population optimum.
The squared standardized residual favors larger scales for difficult predictions, while the logarithmic term charges indiscriminate inflation.

\paragraph{Collapse on acquired views.}
On an acquired angle the push-forward must equal the measurement, even when many volumes remain in $\mathcal{F}$.
\Cref{fig:method}(b) enforces that boundary by presenting an acquired view as a query:
\begin{equation}
\mathcal{L}_{\mathrm{col}}
  = \|\mu(\tilde a)-y_{\tilde a}\|_1 + \mathbb{E}[\sigma(\tilde a)],
  \qquad \tilde a\in\mathcal{A}.
\label{eq:col}
\end{equation}
These two terms are the arrows in \Cref{fig:method}(b): $\mu$ is pulled toward the acquired projection and $\sigma$ toward zero.
The loss is soft, so a finite error and the positive scale floor remain.
We check the resulting scale and mean error numerically rather than asserting exact collapse.
With measurement noise, collapse would describe consistency with the observed sample, not uncertainty about its latent noise-free value. The present gate is interpreted within the simulated measurement protocol.
Acquired-view mean error and scale are checked separately from the unmeasured-view results.
The total loss weights $\mathcal{L}_{\mathrm{NLL}}$ and $\mathcal{L}_{\mathrm{col}}$ equally.
Each specialist runs for $8{,}000$ steps at batch size $1$, from a point-completion initialization, with encoder learning rate zero.

\paragraph{What a calibrated interval means.}
Coverage is the fraction of detector pixels lying in $\mu\pm 1.645\,\sigma$.
Residuals are not Gaussian, so the $\sigma$ that minimizes \eqref{eq:nll} need not hit the nominal $90\%$.
We report untempered scales.
Temperature scaling would need separate calibration data and a joint check against the width charge in \eqref{eq:decompose}.
The static scale in \Cref{fig:idea}(c) is this failure: a scale fit on near residuals pays the miss term on far pixels.

\begin{table}[t]
\centering
\caption{Computing ACUS for one query angle.}
\label{tab:algo}
\scriptsize
\setlength{\tabcolsep}{4pt}
\begin{tabular*}{\columnwidth}{@{}p{\columnwidth}@{}}
\toprule
\textbf{Input:} acquired set $\mathcal{A}$ with views $y_{\mathcal{A}}$, query angle $a'$ with held-out projection $y$, predictor $(\mu,\sigma)$, level $\alpha{=}0.1$. \\
\textbf{Output:} $\mathrm{ACUS}(a')$ and mean scale $u(a')$. \\
\midrule
1: $\delta(a')\leftarrow\min_{a\in\mathcal{A}}|a'-a|$ \\
2: $(\mu,\sigma)\leftarrow\mathrm{Predict}(a';y_{\mathcal{A}})$ using the two nearest views \\
3: $L\leftarrow\mu-1.645\sigma$,\quad $U\leftarrow\mu+1.645\sigma$ \\
4: for each detector pixel $i$ do \\
5: \quad $S_i\leftarrow U_i-L_i$. If $y_i\notin[L_i,U_i]$ then $S_i\leftarrow S_i+\tfrac{2}{\alpha}\,d_i$ \\
6: end for \\
7: $u(a')\leftarrow\mathrm{mean}(\sigma)$. Then $\mathrm{ACUS}(a')\leftarrow\mathrm{mean}(S)$ \\
8: \textbf{gate:} if $a'\in\mathcal{A}$ then require $u(a')\approx 0$ and $\mu\approx y_{a'}$ \\
\bottomrule
\end{tabular*}
\end{table}

\begin{figure*}[t]
\centering
\includegraphics[width=\textwidth]{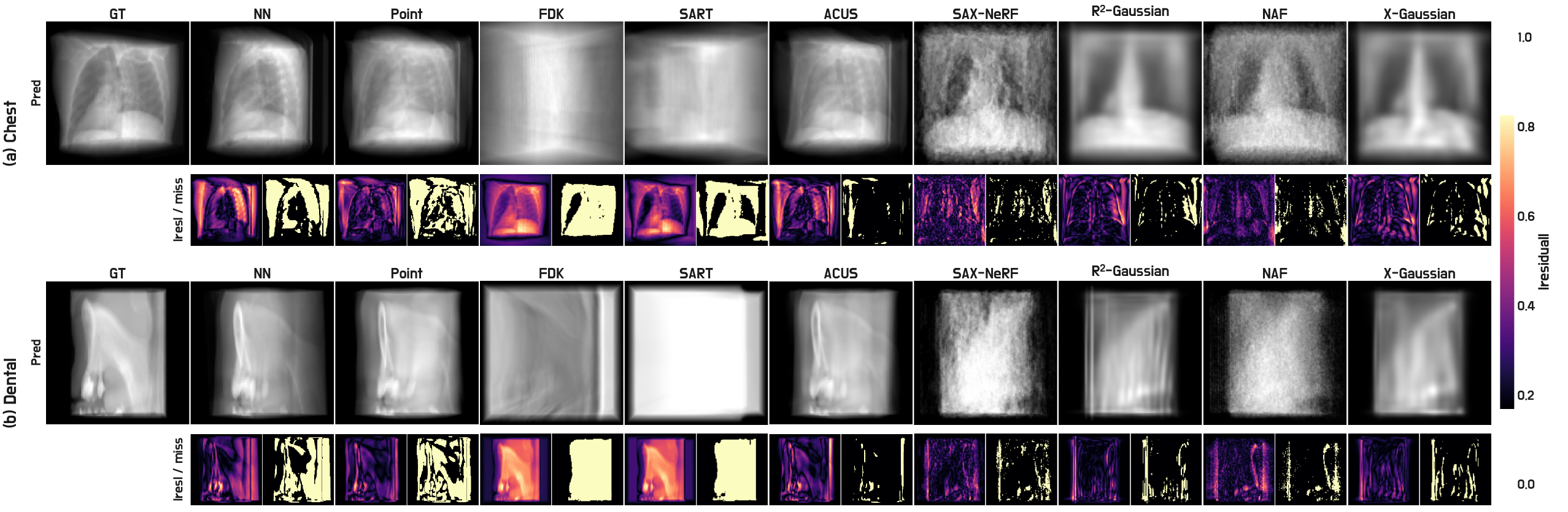}
\caption{Same far-query protocol with classical and learned reconstructors. (a) Chest. (b) Dental.
Each method shows prediction, color residual, and miss mask with a shared colorbar.
Feldkamp-Davis-Kress (FDK)~\cite{feldkamp1984fdk} and Simultaneous Algebraic Reconstruction Technique (SART)~\cite{andersen1984sart} reconstruct the observed orbit and are reprojected at the query angle.
Structure-Aware X-ray Neural Radiodensity Fields (SAX-NeRF), R$^2$-Gaussian, Neural Attenuation Fields (NAF), and X-Gaussian use the same orbit geometry~\cite{cai2024saxnerf,zha2024r2gaussian,zha2022naf,cai2024xgaussian}.}
\label{fig:medext}
\end{figure*}

\section{Experiments}
\label{sec:exp}

\begin{figure}[t]
\centering
\includegraphics[width=\columnwidth]{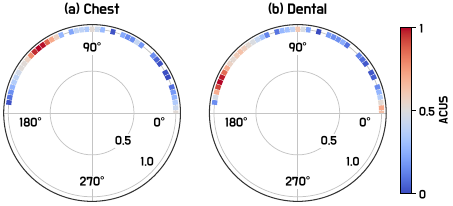}
\caption{Acquisition orbit schematics for chest and dental sparse protocols.
Colored segments mark unobserved views on the 50-view, $180^\circ$ orbit, using the ACUS color scale.}
\label{fig:orbit-schematics}
\end{figure}

\paragraph{Data and protocol.}
Chest slices come from LUNA16~\cite{setio2017luna} and dental volumes from ToothFairy~\cite{bolelli2025toothfairy,cipriano2022tooth}.
Both are rendered onto the same kind of circular cone-beam orbit: 50 views over $180^\circ$.
We train one specialist per anatomy.
Eight cases per domain set the constant residual scale of the copy baselines.
The next 40 cases are the test set.
Every unobserved orbit angle is a query.
Near queries sit less than $20^\circ$ from the closest acquired view. Far queries sit at least $40^\circ$ away.
The main tables use $K{=}8$.
Metrics are mean $\pm$ standard deviation over cases.
\Cref{fig:orbit-schematics} shows the sparse protocol on both anatomies, with unobserved views colored by ACUS.

\begin{figure*}[t]
\centering
\includegraphics[width=0.85\textwidth]{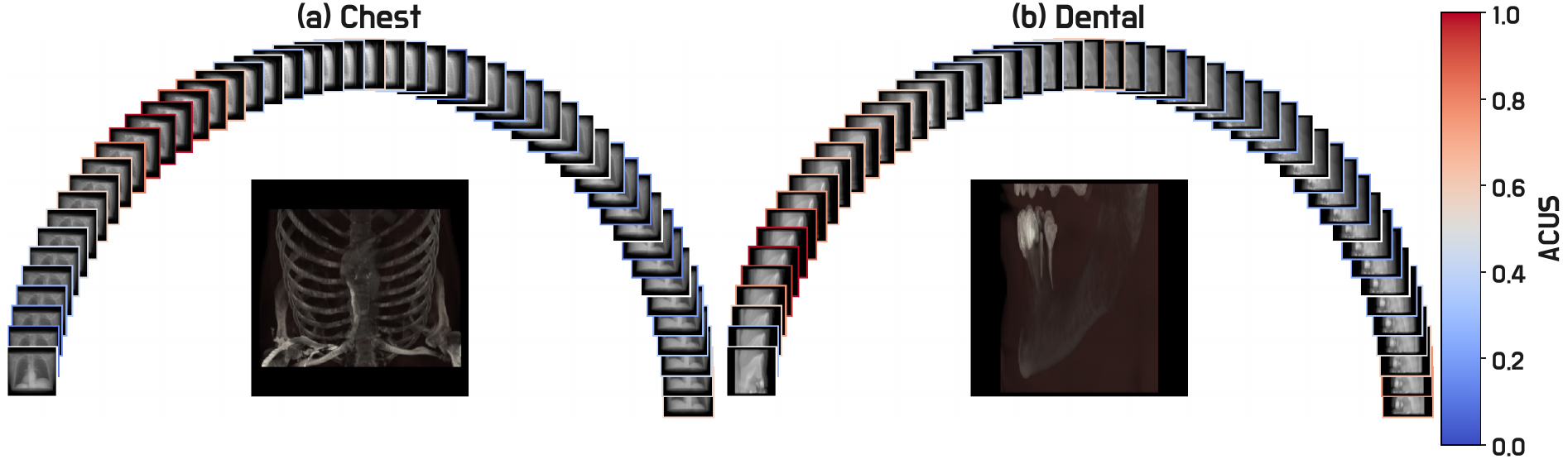}
\caption{Orbit rings with ACUS-colored borders and a central volume rendering.
(a) Chest.
(b) Dental.
Cold borders sit near acquired views.
Warm borders sit on far queries.
The central volume is a spatial anchor, not a reconstruction in the measurement fiber.}
\label{fig:orbit-rings}
\end{figure*}

\paragraph{Baselines.}
Nearest-neighbor and linear angular interpolation copy local measurements.
Point completion is the same backbone trained to match the held-out projection in $\ell_1$.
None of the three predicts a scale.
We give each a single scale equal to its residual on the eight calibration cases, and we use that constant at every angle.
Their acquired-view error is zero by construction, because a query that hits an acquired angle returns the measurement.
The interval model must predict that view.
We therefore report acquired-view width and PSNR only for ACUS, and we do not treat an exact copy as a competing interval.

\begin{figure}[t]
\centering
\includegraphics[width=\columnwidth]{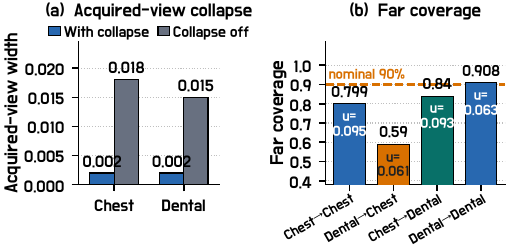}
\caption{Acquired-view collapse and cross-anatomy far coverage on multi-arc queries.
(a) On the 20-case multi-arc ablation split, distinct from the 40-case main tables, removing $\mathcal{L}_{\mathrm{col}}$ raises the acquired scale from $0.002$ to $0.018$ (chest) and $0.015$ (dental). Far-view scales remain $0.095$ (chest) and $0.071$ (dental).
(b) Far coverage for official 40-case same-domain cells and the official transfer evaluation. The dental-to-chest collapse is a negative control, not a transfer success.}
\label{fig:diagnostics}
\end{figure}

\begin{table*}[!t]
\centering
\caption{Official 40-case $K{=}8$ angular bins for ACUS. $u$ is mean scale, not interval length. Cover targets $0.90$. PSNR is of the predictive mean, in dB. Sparse protocols have no $60^\circ$ to $180^\circ$ bin. Multi-arc far columns in \cref{tab:arc} pool the $40^\circ$ to $60^\circ$ and $60^\circ$ to $180^\circ$ bins. The last row has limited case support.}
\label{tab:bins}
\scriptsize
\renewcommand{\arraystretch}{0.84}
\setlength{\tabcolsep}{14.41988pt}
\begin{tabular*}{\textwidth}{lcccccc}
\toprule
\rowcolor{acushead}
& \multicolumn{3}{c}{\textbf{Chest, sparse}} & \multicolumn{3}{c}{\textbf{Dental, sparse}} \\
\rowcolor{acushead}
Gap ($^\circ$) & $u$ & Cover & PSNR & $u$ & Cover & PSNR \\
\midrule
$0$ to $10$ & $0.014{\pm}0.001$ & $0.791{\pm}0.069$ & $33.88{\pm}1.56$ & $0.021{\pm}0.001$ & $0.931{\pm}0.024$ & $33.17{\pm}1.04$ \\
\rowcolor{acusrow}
$10$ to $20$ & $0.027{\pm}0.005$ & $0.848{\pm}0.063$ & $29.49{\pm}2.47$ & $0.030{\pm}0.002$ & $0.911{\pm}0.043$ & $29.15{\pm}1.83$ \\
$20$ to $30$ & $0.055{\pm}0.009$ & $0.851{\pm}0.066$ & $23.11{\pm}1.71$ & $0.047{\pm}0.004$ & $0.900{\pm}0.047$ & $24.71{\pm}1.98$ \\
\rowcolor{acusrow}
$30$ to $40$ & $0.079{\pm}0.004$ & $0.833{\pm}0.087$ & $19.18{\pm}1.50$ & $0.057{\pm}0.001$ & $0.860{\pm}0.051$ & $22.00{\pm}0.57$ \\
$40$ to $60$ & $0.088{\pm}0.003$ & $0.869{\pm}0.052$ & $18.64{\pm}0.99$ & $0.061{\pm}0.002$ & $0.858{\pm}0.083$ & $21.12{\pm}2.04$ \\
\midrule
\rowcolor{acushead}
& \multicolumn{3}{c}{\textbf{Chest, multi-arc}} & \multicolumn{3}{c}{\textbf{Dental, multi-arc}} \\
\rowcolor{acushead}
Gap ($^\circ$) & $u$ & Cover & PSNR & $u$ & Cover & PSNR \\
\midrule
$0$ to $10$ & $0.013{\pm}0.002$ & $0.750{\pm}0.062$ & $32.88{\pm}1.30$ & $0.021{\pm}0.001$ & $0.914{\pm}0.029$ & $32.70{\pm}1.27$ \\
\rowcolor{acusrow}
$10$ to $20$ & $0.034{\pm}0.005$ & $0.785{\pm}0.066$ & $25.75{\pm}1.65$ & $0.034{\pm}0.002$ & $0.853{\pm}0.055$ & $26.38{\pm}1.40$ \\
$20$ to $30$ & $0.057{\pm}0.008$ & $0.859{\pm}0.053$ & $23.15{\pm}1.55$ & $0.047{\pm}0.003$ & $0.862{\pm}0.064$ & $24.16{\pm}1.51$ \\
\rowcolor{acusrow}
$30$ to $40$ & $0.075{\pm}0.012$ & $0.868{\pm}0.062$ & $21.10{\pm}2.31$ & $0.055{\pm}0.003$ & $0.870{\pm}0.067$ & $22.85{\pm}1.68$ \\
$40$ to $60$ & $0.094{\pm}0.008$ & $0.800{\pm}0.054$ & $17.54{\pm}0.53$ & $0.062{\pm}0.002$ & $0.907{\pm}0.025$ & $22.41{\pm}0.81$ \\
\rowcolor{acusrow}
$60$ to $180$ & $0.111{\pm}0.000$ & $0.804{\pm}0.000$ & $16.68{\pm}0.00$ & $0.068{\pm}0.000$ & $0.914{\pm}0.000$ & $21.59{\pm}0.00$ \\
\bottomrule
\end{tabular*}
\end{table*}

\paragraph{Main result.}
On both anatomies and both orbit families, the interval is narrow near an acquired view and several times wider far from one (\cref{tab:sparse,tab:arc,tab:bins}).
On chest sparse orbits the mean scale grows from $0.018$ to $0.088$, and PSNR of the predictive mean falls from $32.48\,\mathrm{dB}$ to $18.64\,\mathrm{dB}$.
Coverage is $0.81$/$0.87$ (chest sparse), $0.77$/$0.80$ (chest multi-arc), $0.92$/$0.86$ (dental sparse), and $0.89$/$0.91$ (dental multi-arc).
Constant-scale baselines have zero reported far coverage, and acquired-view width is $0.002$ in every main cell.
Dental multi-arc point completion is slightly ahead on far PSNR ($22.47$ versus $22.37$ dB) and still has zero reported far coverage.
\Cref{fig:medfar} shows those far queries: constant-scale baselines leave most pixels outside the interval, while ACUS keeps a structured scale and a sparse miss mask.
Near an acquired view the methods look similar, and the break appears in the far columns.
\Cref{fig:medext} repeats the far query for FDK~\cite{feldkamp1984fdk} and SART~\cite{andersen1984sart}, reprojected from the observed orbit, and for SAX-NeRF, R$^2$-Gaussian, NAF, and X-Gaussian~\cite{cai2024saxnerf,zha2024r2gaussian,zha2022naf,cai2024xgaussian}.
Analytical reprojections lose structure.
The learned images stay sharp, and one calibration scale makes their miss masks follow the copy baselines.
Supplement, Sec.~D gives the near/far grids, and Secs.~E to H collect further boards.
On chest sparse orbits the copy baselines use constant scales $0.014$, $0.012$, and $0.013$, with near coverage $0.61$, $0.70$, and $0.72$.
ACUS uses $0.018$ near an acquired view and $0.088$ on far queries, and its far PSNR is $18.64\,\mathrm{dB}$ against $16.40$, $16.34$, and $14.14\,\mathrm{dB}$.
Dental sparse orbits repeat that split at a smaller far scale, from $0.024$ to $0.061$, with far PSNR $21.12\,\mathrm{dB}$ against $18.51$, $18.54$, and $20.95\,\mathrm{dB}$.
Under multi-arc packing the chest near scales already agree ($0.022$, $0.022$, and $0.026$ against $0.023$), while the far scale opens to $0.095$ and far PSNR is $17.48\,\mathrm{dB}$ against $15.51$, $15.28$, and $13.59\,\mathrm{dB}$.
Dental multi-arc scales are $0.027/0.063$ at coverage $0.89/0.91$.
Case-level scale deviations on these ACUS rows are at most $0.003$, except $0.008$ on chest multi-arc far queries.

\begin{figure}[t]
\centering
\includegraphics[width=0.72\columnwidth]{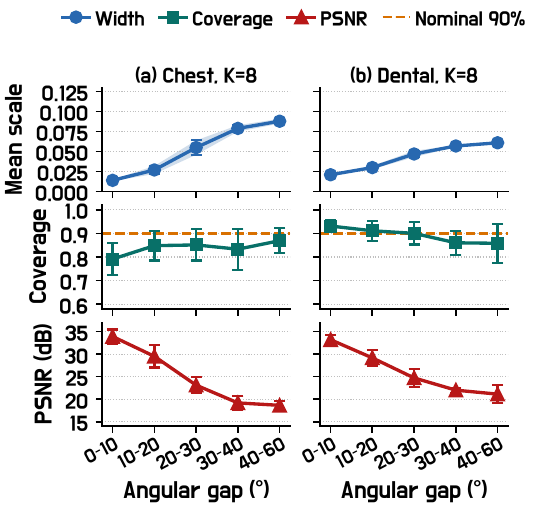}
\caption{Sparse $K{=}8$ gap bins on the official 40-case split.
(a) Chest. (b) Dental.
Curves show mean scale, empirical coverage, and PSNR of the predictive mean.
The $40^\circ$ to $60^\circ$ bin is the far column of \cref{tab:sparse}.
Error bars are case-level standard deviations, and the dashed line is the nominal $90\%$ rate.
Mean ACUS is omitted because the outside-distance term is unreported.}
\label{fig:gap}
\end{figure}

\Cref{fig:gap} and \cref{tab:bins} split the official sparse set into angular bins.
On these sparse bins, coverage stays between $0.79$ and $0.87$ on chest and between $0.86$ and $0.93$ on dental.
On chest sparse orbits the mean scale steps through $0.014$, $0.027$, $0.055$, $0.079$, and $0.088$, and PSNR falls from $33.88\,\mathrm{dB}$ to $18.64\,\mathrm{dB}$, with the largest step from $29.49\,\mathrm{dB}$ to $23.11\,\mathrm{dB}$ between the $10^\circ$ to $20^\circ$ and $20^\circ$ to $30^\circ$ bins.
Dental sparse scales rise more slowly, through $0.021$, $0.030$, $0.047$, $0.057$, and $0.061$.
The multi-arc bins are separate: chest coverage there is $0.750$ in the $0^\circ$ to $10^\circ$ row, and the $60^\circ$ to $180^\circ$ row has limited case support.
Chest multi-arc coverage is $0.800$ in the $40^\circ$ to $60^\circ$ bin, below $0.859$ and $0.868$ in the two middle bins.
That tail row is $u{=}0.111$, coverage $0.804$, and $16.68\,\mathrm{dB}$ on chest, and $u{=}0.068$, coverage $0.914$, and $21.59\,\mathrm{dB}$ on dental.
The pooled multi-arc far cells sit next to the $40^\circ$ to $60^\circ$ rows rather than on that thin tail (Supplement, Sec.~C).
Supplement, Sec.~C also shows a $32.4^\circ$ case from the $30^\circ$ to $40^\circ$ bin, which is not a far query.
\Cref{fig:orbit-rings} draws the same score on the acquisition ring.
The central volume is a spatial anchor.

\paragraph{Collapse ablation.}
Removing $\mathcal{L}_{\mathrm{col}}$ on a 20-case multi-arc split, separate from the 40-case tables, leaves far width unchanged at $0.095$ (chest) and $0.071$ (dental) and raises acquired-view width from $0.002$ to $0.018$ (chest) and $0.015$ (dental) (\cref{fig:diagnostics}(a)).
The term is tied to acquired views.
The dental far width on this split is not the main-table value $0.063$.
Far coverage on the same split moves from $0.80$ to $0.79$ on chest and from $0.90$ to $0.89$ on dental.
On the official cells, predictive PSNR at acquired angles is $46.4$ and $52.6\,\mathrm{dB}$ on chest and $49.5$ and $54.9\,\mathrm{dB}$ on dental, for sparse and multi-arc orbits, at the same acquired width $0.002$.
Supplement, Sec.~J gives the ablation rows and the acquired-view PSNR on all four official cells.

\paragraph{Where the width is anatomy, not geometry.}
A chest model on dental multi-arc scans keeps a wide far interval ($0.093{\pm}0.013$), coverage $0.84$, and far PSNR $19.59{\pm}2.08$.
A dental model on chest scans keeps far width $0.061{\pm}0.001$, coverage $0.59$, and far PSNR $16.58{\pm}1.12$, against about $0.80$ to $0.87$ in domain (\cref{fig:diagnostics}(b)).
Dental-to-chest is the negative control: the narrow interval is an anatomical prior.
Supplement, Sec.~G shows the corresponding transfer panels.

\paragraph{Long-arc stress test.}
On 16 cases with one arc of $50^\circ$ to $120^\circ$, far width stays large ($0.090$ chest, $0.064$ dental; near/far PSNR $32.45/18.68$ and $31.25/21.28$; far coverage $0.82$ and $0.88$).
Near width is $0.027$ on both anatomies, against far widths $0.090$ and $0.064$.
The check uses another acquisition family and is not a matched 40-case estimate (Supplement, Sec.~J).

\begin{figure}[t]
\centering
\includegraphics[width=\columnwidth]{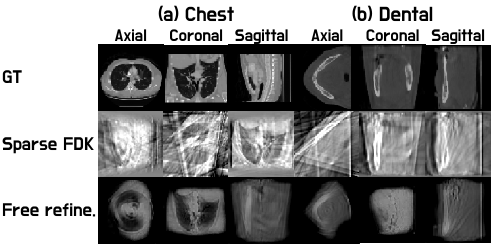}
\caption{Official $64^3$ slices side by side in one column, chest on the left and dental on the right.
(a) Chest.
(b) Dental.
Columns are axial, coronal, and sagittal.
Rows are the ground-truth volume, sparse FDK, and a free residual refinement that can look closer to the anatomy while agreeing less with the acquired projections, and does not yield a volume in $\mathcal{F}$ (\Cref{sec:volume}).}
\label{fig:volume-slices}
\end{figure}

\paragraph{Volume readout, not a result.}
\label{sec:volume}
On 12 sparse cases at $64^3$, \Cref{fig:volume-slices} compares a free residual refinement with sparse FDK and the ground-truth volume.
Volume PSNR rises (chest $14.04{\pm}1.11$ vs.\ $6.82{\pm}0.36$, dental $18.26{\pm}0.82$ vs.\ $9.41{\pm}0.56$) and agreement with the acquired projections falls (chest $9.79{\pm}1.10$ vs.\ $13.32{\pm}1.41$, dental $9.13{\pm}0.87$ vs.\ $10.41{\pm}0.99$).
The refinement was not trained with query-angle likelihood, and the FDK intensities are not on the measurement scale.
Only $0.07$ to $0.15$ of the query-angle forward projections fall inside the predicted intervals, so the diagnostic does not yield a volume in $\mathcal{F}$.
Supplement, Sec.~H pairs these slices with the reprojection curves.

\section{Discussion}
\label{sec:disc}

The tables separate point-estimate accuracy from interval reliability.
Near acquired support, accurate local baselines coexist with a small learned scale, while far away the constant scales have zero reported coverage and the learned scales open.
Dental multi-arc point completion leads slightly on far PSNR and still misses the far interval.
One residual scale leaves angle-conditioned calibration, a trained ensemble, and a conformal interval on the same outputs untested, and acquired-view collapse leaves far coverage unchecked.
Chest coverage stays below the nominal rate, so further calibration is required before the $90\%$ level is an operating guarantee.
Correlated misses can sit on a small structure and barely move the detector average, so pixelwise coverage and the diagonal Gaussian fall short of a simultaneous confidence region for a projection or a volume.

On the far-query boards, FDK and SART lose structure, while SAX-NeRF, R$^2$-Gaussian, NAF, and X-Gaussian stay sharp, and one calibration scale makes their miss masks follow the copy baselines (\cref{fig:medext}).
Under \eqref{eq:decompose}, a covered query scores $3.29\,u$ and every miss adds twenty times its outside distance.
Learned rows omit that distance, so zero far coverage activates the penalty without an ACUS rank.
A scalar is printed only for the analytical reprojectors (Supplement, Sec.~J); Sec.~E splits it into width and miss.

On 12 sparse cases, only $0.07$ to $0.15$ of the query-angle forward projections fall inside the predicted intervals (\cref{fig:volume-slices}).
Volume similarity can rise while the estimate leaves $\mathcal{F}$.

On the shared $50$-view, $180^\circ$ trajectory, a chest model on dental scans keeps coverage $0.84$ with a wide far interval, while a dental model on chest scans keeps a far width near $0.061$ and coverage $0.59$.
The opened scale follows the training anatomy.
Pooled chest coverage of $0.77$ to $0.80$ on multi-arc queries is the in-domain counterpart, while the $0^\circ$ to $10^\circ$ bin in \cref{tab:bins} is $0.750$.

The present evidence supports a practical audit of gap-conditioned scale and coverage, the acquired boundary, and the miss magnitude when an aggregate interval score is published.

\section{Conclusion}
\label{sec:limit}

We present \textsc{ACUS}, an angular calibrated uncertainty score for incomplete-view CT. \textsc{ACUS} assigns a nominal predictive interval on held-out projections by distance to the nearest acquired view, with acquired-view collapse as a necessary consistency check. A feed-forward mean and scale trained with held-out likelihood gate wider intervals at larger gaps. \textsc{ACUS} uses a diagonal Gaussian on detector pixels and is evaluated on simulated orbits. Cross-anatomy transfer is a negative control, not a domain-generalization claim. Constant-scale point predictors fail far bins despite strong near-angle PSNR.

\section*{LLM Usage Statement}
ChatGPT assisted with manuscript editing, LaTeX formatting, figure typography, and checks of references. The authors take responsibility for the content and results of this work.

{
    \small
    \bibliographystyle{ieeenat_fullname}
    \bibliography{main}
}

\clearpage
\appendix
\setcounter{dbltopnumber}{2}

\begin{center}
{\Large\bfseries Supplementary Material}
\end{center}
\vspace{0.75em}
\phantomsection
\label{sec:supp-front}

\noindent This supplement reprints the official angular bins, the collapse ablation, the cross-anatomy transfer, the long-arc stress test, the analytical reprojections, and the $64^3$ volume readout.
It also records the comparisons that the pooled cells hide.
Interval scale grows with the gap on every official split.
Headline coverage lies between $77\%$ and $92\%$.
A single residual scale misses far-angle pixels, and a higher volume PSNR can leave the measurement fiber.
\Cref{fig:supp-official-curves,fig:supp-binned} are drawn from those tables and from one later chest rerun that is not substituted for the official cells.
Case identifiers are removed from every figure.

\section{Broader impact and asset notes}
\label{sec:supp-impact}

ACUS is an interval-evaluation tool for incomplete-view computed tomography (CT) protocols.
It scores a finished acquisition.
It is not an acquisition policy, a stopping rule, or a certificate that the mean image is the only volume consistent with the measurements.
\Cref{sec:supp-3d} is the concrete form of that limit: volume PSNR can rise while agreement with the acquired projections falls, and only $0.07$ to $0.15$ of the query-angle forward projections fall inside the predicted intervals.

\paragraph{Assets.}
Chest slices come from LUNA16~\cite{setio2017luna} and dental volumes from ToothFairy~\cite{bolelli2025toothfairy,cipriano2022tooth} under their published licenses.
All orbits are simulated cone-beam renderings on one 50-view, $180^\circ$ geometry, so a transfer experiment changes the object and the specialist, not the angular grid.
No patient-identifying metadata appears in the figures.
A single displayed case is evidence for the 40-case, 20-case, 16-case, or 12-case table named in its caption.

\section{Protocol details beyond the main text}
\label{sec:supp-protocol}

\paragraph{Shared geometry.}
Both domains are rendered onto a circular cone-beam orbit of 50 views spanning $180^\circ$.
Every query is one of those rendered views that the mask left out, so the evaluation never leaves the simulated grid.
The gap $\delta(a')$ is the minimum absolute distance on this sampled trajectory from the query to a view that the mask kept.
Endpoints of the $180^\circ$ arc are not glued together.
A query near one end is far from a view near the other end even though both sit on the same semicircle.
This matches the definition in the main text and is why the multi-arc tables contain a $60^\circ$ to $180^\circ$ bin that the sparse tables do not.
Sparse sampling spreads eight views around the semicircle, so the largest hole stays inside $40^\circ$ to $60^\circ$.
Packing the same eight views into two or three short arcs opens at least one much larger hole, and that hole is the last row of \cref{tab:supp-chest-arc-bins,tab:supp-dental-arc-bins}.

\paragraph{Why the two headline families share a view budget.}
Sparse $K{=}8$ picks eight angles on the 50-view circle.
Multi-arc $K{=}8$ packs the same budget into two or three short arcs of $25^\circ$ to $45^\circ$.
Holding $K$ fixed keeps angular arrangement from being confounded with the number of measurements.
The two families are therefore reported separately at the same $\delta$, including in the supplement tables.

\paragraph{Near, far, and the bins underneath them.}
Near queries sit under $20^\circ$ from the closest acquired view.
Far queries sit at least $40^\circ$ away.
Angles from $20^\circ$ to $40^\circ$ are scored and are excluded from both headline columns.
\Cref{fig:midgap} is one such case, at $32.4^\circ$.
Reported means average per-case query means, and the standard deviations are case-level.
Neighboring detector pixels share anatomy, so these deviations are not pixelwise standard errors.

\paragraph{Calibration of copy baselines.}
Nearest-neighbor, linear interpolation, and point completion receive one residual scale fit on eight calibration cases per domain.
Those eight cases are disjoint from the 40-case test split.
The fitted scale is one positive number per method and domain, held fixed at every query angle.
This is the constant-scale failure the miss penalty charges.
It is not an angle-conditioned calibrator, an ensemble, or a conformal interval on the same outputs.
Acquired-view error of a copy baseline is zero by construction, because a query that lands on an acquired angle returns the measurement.
The interval model has to predict that view from its neighbors.
We therefore report acquired-view width and predicted PSNR only for ACUS.
An exact copy is not entered as a competing interval in \cref{tab:supp-acq-width}.

\paragraph{External reconstructors.}
Structure-Aware X-ray Neural Radiodensity Fields (SAX-NeRF)~\cite{cai2024saxnerf}, R$^2$-Gaussian~\cite{zha2024r2gaussian}, Neural Attenuation Fields (NAF)~\cite{zha2022naf}, and X-Gaussian~\cite{cai2024xgaussian} are evaluated on the same far-query geometry used for the ACUS grids.
Feldkamp-Davis-Kress (FDK)~\cite{feldkamp1984fdk}, the Simultaneous Algebraic Reconstruction Technique (SART)~\cite{andersen1984sart}, and adaptive-steepest-descent projection onto convex sets (ASD-POCS)~\cite{sidky2008asdpocs} reconstruct the observed orbit and are reprojected at the query.
Each method returns a point image.
We attach the same residual visualization used for the copy baselines so that miss masks are comparable across columns.
These methods were trained or derived as point reconstructors, not as interval estimators.
The scalar ACUS printed in \cref{tab:supp-analytical} is computed only for the three analytical reprojectors, which is the statement in the main discussion.
Learned reconstructors appear as images and miss masks.
They are not given a second, specially tuned scale in this supplement.

\section{Gap-conditioned evaluation}
\label{sec:supp-curves}

\begin{figure*}[t]
\centering
\includegraphics[width=\textwidth]{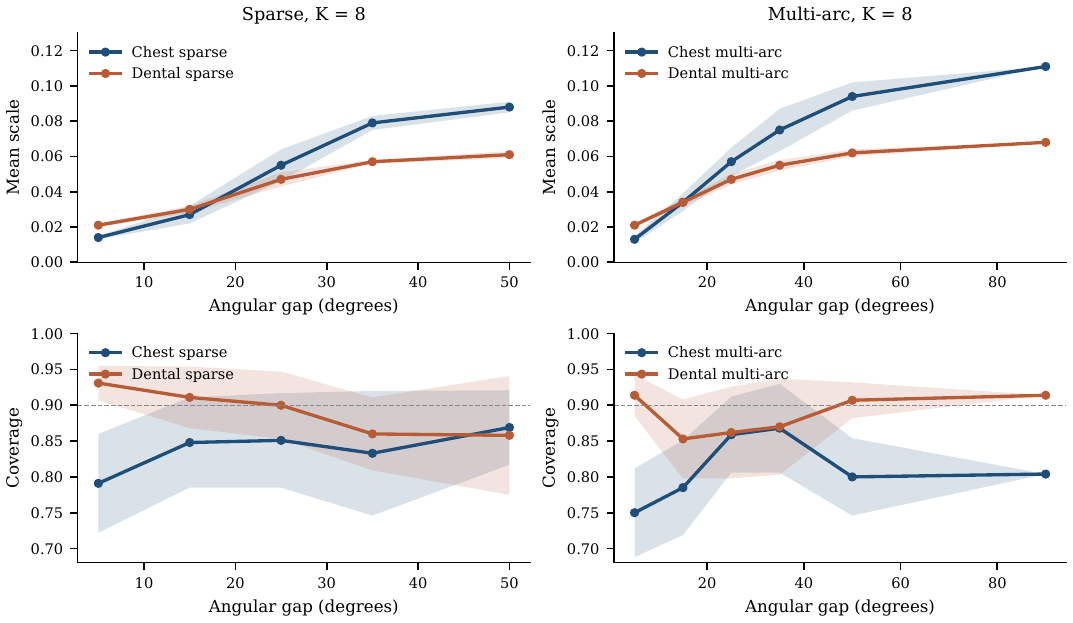}
\caption{Official $K{=}8$ bins, 40 cases per domain.
Bands are case-level standard deviations.
The dashed line is the nominal $90\%$ rate.
The $60^\circ$ to $180^\circ$ chest and dental multi-arc points have no case-level deviation because that bin contains one query pattern in the official split.
These curves use the printed supplement tables, not a later rerun.}
\label{fig:supp-official-curves}
\end{figure*}

The official 40-case bins also appear in the main-text experiments.
The same values are repeated here in \cref{tab:supp-chest-sparse-bins,tab:supp-chest-arc-bins,tab:supp-dental-sparse-bins,tab:supp-dental-arc-bins}.
\Cref{fig:supp-official-curves} plots those four tables together.
The mechanism the main paper claims is visible without pooling.
Mean scale rises monotonically with the gap on every official split: chest sparse from $0.014$ to $0.088$, chest multi-arc from $0.013$ to $0.111$, dental sparse from $0.021$ to $0.061$, and dental multi-arc from $0.021$ to $0.068$.
Coverage does not follow that rise.
It stays inside $0.75$ to $0.93$, below the nominal $90\%$ rate on chest and closer to it on dental.
The point-estimate PSNR falls over the same bins, which is the separation ACUS is built to report: a wider interval can remain approximately calibrated while the mean image becomes a poorer point estimate.
A static scale cannot trace this curve, because it has one magnitude at every gap.

\begin{figure*}[t]
\centering
\begin{tabular}{@{}c@{\hspace{0.35em}}c@{}}
\includegraphics[width=0.48\textwidth]{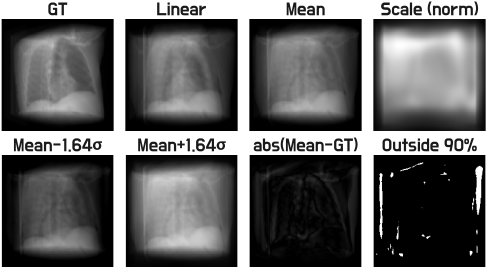} &
\includegraphics[width=0.48\textwidth]{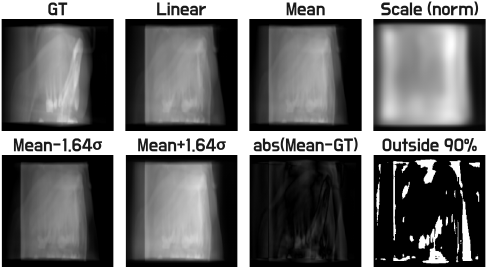} \\[-0.15em]
{\footnotesize \textbf{(a)} Chest sparse, $32.4^\circ$ gap} &
{\footnotesize \textbf{(b)} Dental sparse, $32.4^\circ$ gap}
\end{tabular}
\caption{Held-out $32.4^\circ$ mid-range intervals in the $30^\circ$ to $40^\circ$ bin, not a far query of at least $40^\circ$.
Panels show ground truth (GT), linear interpolation, the ACUS mean, a scale map, the $90\%$ bounds, the residual, and outside-interval pixels.}
\label{fig:midgap}
\end{figure*}

\Cref{fig:midgap} shows one chest case and one dental case at a mid-range gap of $32.4^\circ$, which lies in the $30^\circ$ to $40^\circ$ bin and is not a far query of at least $40^\circ$.
The columns follow the same convention as the main far-query boards: ground truth, linear interpolation, the ACUS mean, the scale map, the published $90\%$ bounds, the residual, and the miss mask.
The scale maps are already structured at this mid-range gap, and the miss masks are not uniform across the detector.

\paragraph{How the sparse bins sit under the headline cells.}
On chest sparse orbits, scale rises from $0.014$ in the $0^\circ$--$10^\circ$ bin to $0.088$ in the $40^\circ$--$60^\circ$ bin (\cref{tab:supp-chest-sparse-bins}).
The rise is monotone across the five bins: $0.014$, $0.027$, $0.055$, $0.079$, $0.088$.
The largest single step is from the $10^\circ$--$20^\circ$ bin to the $20^\circ$--$30^\circ$ bin, where the mean scale roughly doubles and PSNR of the predictive mean falls from $29.49\,\mathrm{dB}$ to $23.11\,\mathrm{dB}$.
That is the same transition the protocol uses to leave the near group.
The pooled near cell in main \cref{tab:sparse} is $u{=}0.018$, coverage $0.81$, PSNR $32.48\,\mathrm{dB}$.
It mixes the $0^\circ$--$10^\circ$ bin ($u{=}0.014$, coverage $0.791$, PSNR $33.88\,\mathrm{dB}$) with the $10^\circ$--$20^\circ$ bin ($u{=}0.027$, coverage $0.848$, PSNR $29.49\,\mathrm{dB}$).
The pooled cell and the first bin are different summaries.
The far cell $u{=}0.088$, coverage $0.87$, PSNR $18.64\,\mathrm{dB}$ matches the $40^\circ$--$60^\circ$ row exactly, because sparse orbits have no wider bin.
Dental sparse scale rises more slowly and saturates: $0.021$, $0.030$, $0.047$, $0.057$, $0.061$ (\cref{tab:supp-dental-sparse-bins}).
The far cell $0.061$ is again the last bin.
The pooled dental near cell $0.024$ likewise sits between the first bin $0.021$ and the second bin $0.030$.
Far PSNR remains higher than on chest ($21.12\,\mathrm{dB}$ against $18.64\,\mathrm{dB}$) at a narrower scale ($0.061$ against $0.088$).
The main conclusion that the opened scale follows the training anatomy is already visible inside these two in-domain curves, before the transfer table is consulted.

\paragraph{Coverage does not track scale.}
Chest sparse coverage across the five bins is $0.791$, $0.848$, $0.851$, $0.833$, $0.869$.
It stays inside the band the main text quotes for the sparse chest curve, between $0.79$ and $0.87$, and it is not monotone in the gap.
The $30^\circ$--$40^\circ$ bin is the local dip ($0.833{\pm}0.087$), and it is also the bin with the widest case-level spread in that table.
Dental sparse coverage is $0.931$, $0.911$, $0.900$, $0.860$, $0.858$.
It starts near the nominal rate and drifts down to the mid-$0.80$s, which is why the headline dental sparse pair can be $0.92$ near and $0.86$ far without either number being a summary of every bin.
Under main \eqref{eq:decompose} a fully covered detector scores $3.29\,u$, and each missed pixel adds twenty times its outside distance.
The mean outside distance is unreported, so these rows cannot be ranked by a scalar ACUS.

\paragraph{Multi-arc bins and the thin last row.}
Chest multi-arc scale continues the same monotone opening, and it does not stop at $60^\circ$: $0.013$, $0.034$, $0.057$, $0.075$, $0.094$, $0.111$ (\cref{tab:supp-chest-arc-bins}).
The headline far cell $0.095$ pools the $40^\circ$--$60^\circ$ row ($0.094$, coverage $0.800$, PSNR $17.54\,\mathrm{dB}$) with the $60^\circ$--$180^\circ$ row ($0.111$, coverage $0.804$, PSNR $16.68\,\mathrm{dB}$).
An equal mixture of those two rows would sit near $0.102$.
The published pool sits at $0.095$, next to the $40^\circ$--$60^\circ$ row, so the thin last bin does not dominate the headline far cell even though its own scale is the widest in the paper.
Dental multi-arc scale is $0.021$, $0.034$, $0.047$, $0.055$, $0.062$, $0.068$, and the headline far cell $0.063$ is the corresponding pool of $0.062$ and $0.068$ (\cref{tab:supp-dental-arc-bins}).
Dental coverage on those two far bins is $0.907$ and $0.914$, which is why the pooled far coverage in main \cref{tab:arc} can sit at $0.91$ even though several nearer dental bins are in the mid-$0.80$s.
The last row prints a case-level standard deviation of $0.000$ because support is too thin to show scatter.
That zero is not a population variance, and $16.68\,\mathrm{dB}$ or $21.59\,\mathrm{dB}$ is not a high-precision estimate.
The chest multi-arc $0^\circ$--$10^\circ$ bin is the other pooled quantity that needs the row underneath it: coverage $0.750{\pm}0.062$ at scale $0.013$, against a headline near cell of coverage $0.77$ at scale $0.023$.
The far cell is $0.80$.
Undercoverage relative to $0.90$ is present on both sides of the gap, and the lowest rate is the nearest bin, where PSNR is highest ($32.88\,\mathrm{dB}$).

The $32.4^\circ$ case belongs to the $30^\circ$--$40^\circ$ row: chest sparse scale $0.079$, coverage $0.833$, PSNR $19.18\,\mathrm{dB}$; dental sparse scale $0.057$, coverage $0.860$, PSNR $22.00\,\mathrm{dB}$.
Those means sit between the near scales near $0.02$ and the chest far scale $0.088$.
The panel should be cited with that row, not with the far column of main \cref{tab:sparse}.

\begin{figure}[t]
\centering
\includegraphics[width=0.6\columnwidth]{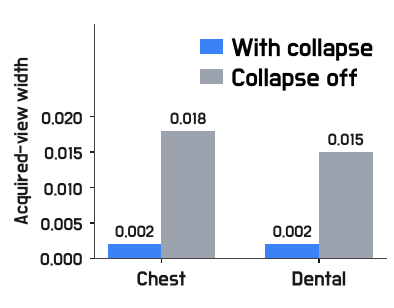}
\caption{Acquired-view collapse on the 20-case multi-arc ablation split. The panel is the same acquired-view collapse diagnostic as main \cref{fig:diagnostics}(a). The values agree with \cref{tab:supp-collapse}.}
\label{fig:supp-winkler}
\end{figure}

\Cref{fig:supp-winkler} scores acquired angles on the 20-case multi-arc ablation, not the unobserved angles in the bins above.
With $\mathcal{L}_{\mathrm{col}}$, acquired width is $0.002$.
Without it, acquired width rises to $0.018$ (chest) and $0.015$ (dental), while far width stays at $0.095$ and $0.071$ (\cref{tab:supp-collapse}).
Chest far width $0.095$ matches the 40-case multi-arc cell.
Dental far width $0.071$ does not match the 40-case cell $0.063$.
The two splits stay separate.

\section{Near versus far medical grids}
\label{sec:supp-nearfar}

\begin{figure*}[t]
\centering
\includegraphics[width=\textwidth]{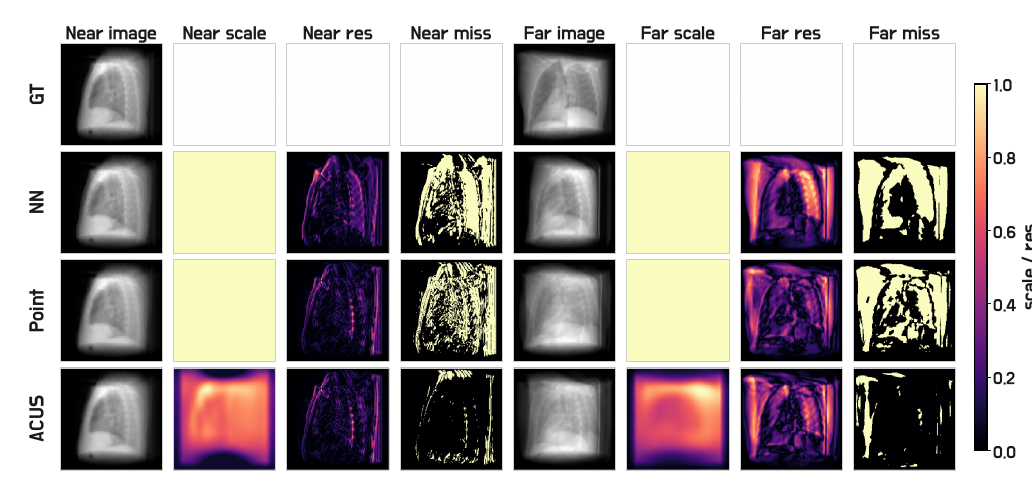}
\caption{Near versus far readout on one chest sparse case.
Left block: near query.
Right block: far query.
Color panels show scale and absolute residual with a shared colorbar.
Near the acquired view every method looks similar.
Far away only ACUS reports a non-constant scale and keeps most pixels inside the interval.}
\label{fig:supp-nearfar-chest}
\end{figure*}

\begin{figure*}[t]
\centering
\includegraphics[width=\textwidth]{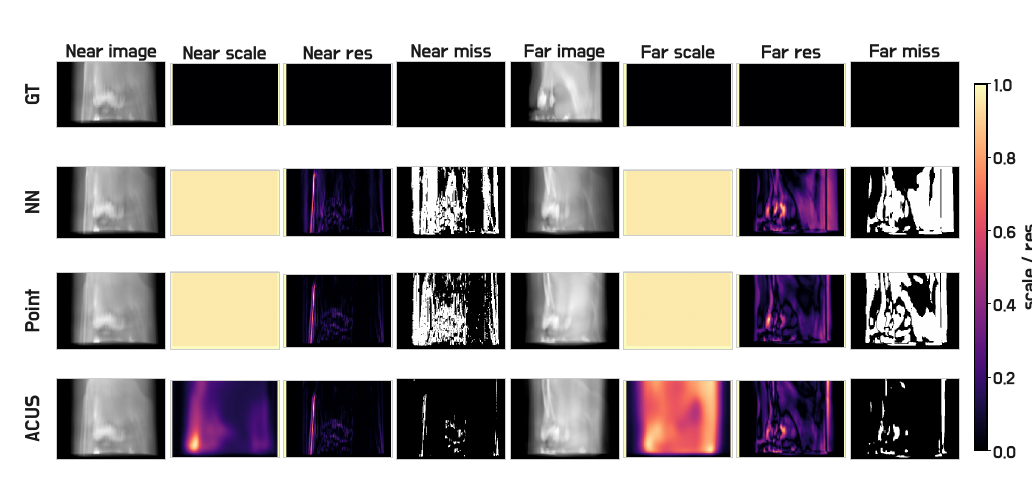}
\caption{Dental near versus far with color scale and residual maps and a shared colorbar.
The dental specialist keeps a narrower absolute scale than the chest specialist, while far coverage remains near the nominal rate on its own domain.}
\label{fig:supp-nearfar-dental}
\end{figure*}

On a chest sparse near query the four means span $2.2\,\mathrm{dB}$: $30.27$, $32.34$, $32.17$, and $32.48$ for nearest, linear, point completion, and ACUS.
The far block is where they separate.
The chest sparse constants are $0.014$, $0.012$, and $0.013$, the size of the $0^\circ$--$10^\circ$ ACUS bin, against an ACUS far scale of $0.088$.
Main \cref{tab:sparse} records zero far coverage for every constant-scale row.
In \cref{fig:supp-nearfar-chest} the far miss masks of those rows fill the detector, and the ACUS miss mask stays sparse.
A full mask means the factor-of-$20$ term in main \eqref{eq:decompose} is active on nearly every pixel.
The figure does not supply the outside distance, so it supports the coverage column and not a scalar ranking.

\paragraph{Dental keeps a narrower absolute scale.}
\Cref{fig:supp-nearfar-dental} uses the same layout.
Dental sparse far scale is $0.061$ with coverage $0.86$; chest sparse far scale is $0.088$ with coverage $0.87$.
The cooler dental map is a smaller absolute scale at similar far coverage, not a better-calibrated interval.
Dental near coverage is $0.92$ and chest near coverage is $0.81$, so the two maps also have to be read with their own near rates.
The same dental scale, evaluated on chest in \cref{tab:supp-transfer}, drops far coverage to $0.59$.

\begin{figure*}[t]
\centering
\includegraphics[width=\textwidth]{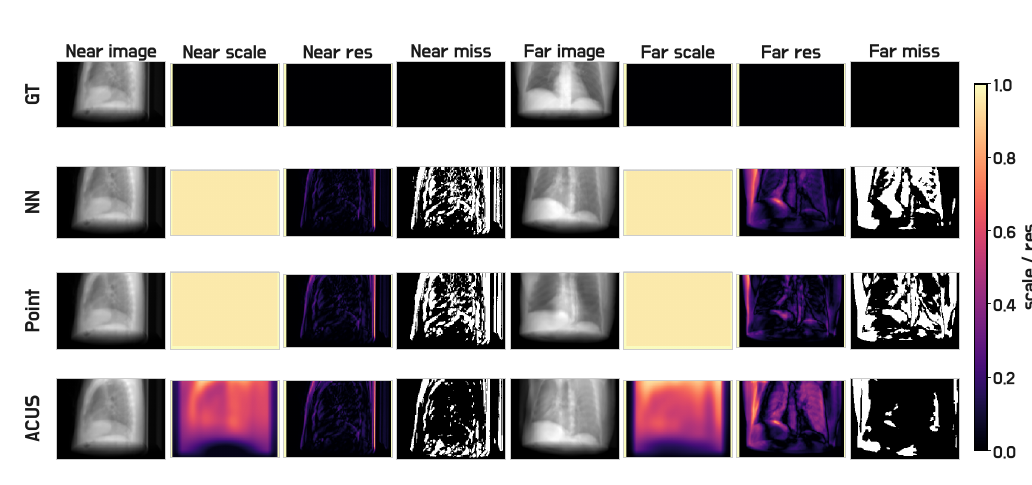}
\caption{Chest multi-arc near versus far with color residuals.
Multi-arc packing changes which angles are local, but the same near-narrow and far-wide pattern appears.}
\label{fig:supp-nearfar-multi}
\end{figure*}

\paragraph{Multi-arc packing moves which angles are local.}
\Cref{fig:supp-nearfar-multi} is the chest multi-arc version of the same near-versus-far readout.
Packing eight views into two or three short arcs changes the set of queries that fall under $20^\circ$.
Inside an arc, neighboring acquired views make a query local even though most of the semicircle is empty.
Outside the arcs, a query can sit $60^\circ$ or more from every measurement, which sparse $K{=}8$ does not do.
The panel still shows the near-narrow and far-wide pattern, at chest multi-arc scale $0.023/0.095$, coverage $0.77/0.80$, and PSNR $29.59/17.48$.
Near PSNR is $2.9\,\mathrm{dB}$ below the sparse near value $32.48\,\mathrm{dB}$: a multi-arc near query is local only to a short arc.
The near scale stays small ($0.023$, next to the sparse near scale $0.018$) while the far scale opens to $0.095$.
Sparse and multi-arc rows are not averaged, because their near images come from different neighborhoods.

\section{External methods and residual mosaics}
\label{sec:supp-external}

\begin{figure*}[t]
\centering
\includegraphics[width=\textwidth]{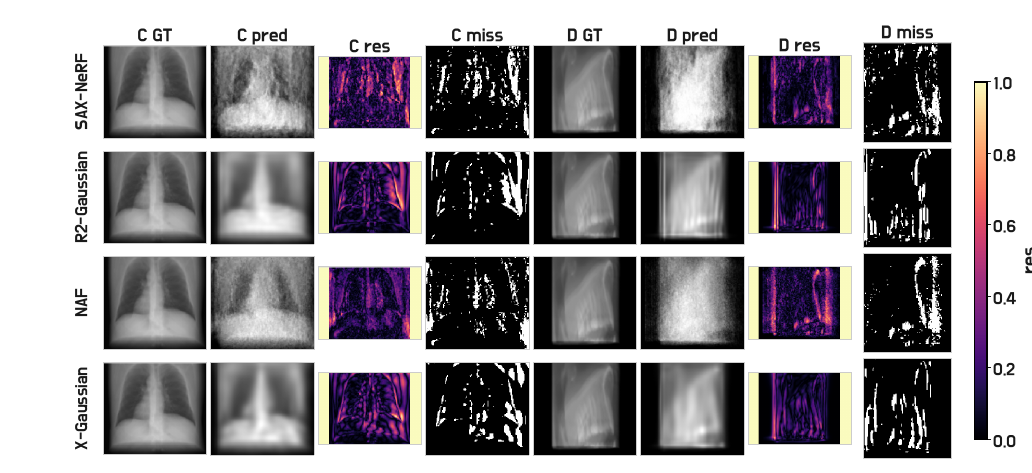}
\caption{External learned methods with color residuals on chest and dental far queries.
Each row is one reconstructor.
Columns alternate prediction, residual, and miss for chest and dental.}
\label{fig:supp-external}
\end{figure*}

\Cref{fig:supp-external} separates prediction, residual, and miss for the learned reconstructors on both anatomies.
The analytical rows are a same-geometry point baseline, not a matched rerun of the official ACUS cells.
On \cref{tab:supp-analytical}, FDK reaches $11.13/11.60\,\mathrm{dB}$ with coverage $0.57/0.50$ and ACUS $0.739$.
SART reaches $15.58/14.51\,\mathrm{dB}$ with coverage $0.60/0.50$ and ACUS $0.509$.
ASD-POCS matches SART at the displayed precision ($15.57/14.51$, coverage $0.60/0.50$, ACUS $0.509$), so the convex-set step does not separate from SART here.
Far PSNR of $11.6$ to $14.5\,\mathrm{dB}$ and far coverage $0.50$ both sit below the official chest ACUS cells ($17.48$ to $18.64\,\mathrm{dB}$, far coverage $0.80$ to $0.87$).

\paragraph{Reading the printed scalar.}
These three rows are the only place a scalar ACUS is attached to a point reconstructor.
Main \eqref{eq:decompose} splits that scalar into width and outside distance, $\mathrm{ACUS}=3.29\,u+20\,\bar d$.
For FDK, $3.29\times 0.144\approx 0.47$, so a score of $0.739$ leaves about $0.26$ to the miss term.
For SART, $3.29\times 0.086\approx 0.28$, and a score of $0.509$ leaves about $0.23$ to the miss term.
The miss terms are similar, about $0.26$ and $0.23$, so FDK's larger width raises the scalar while far coverage stays at $0.50$.
The learned rows do not print a scalar, so the same split cannot be run on SAX-NeRF or on point completion.
Zero far coverage says the miss term is active and does not say how large the outside distances are.

\begin{figure*}[t]
\centering
\includegraphics[width=\textwidth]{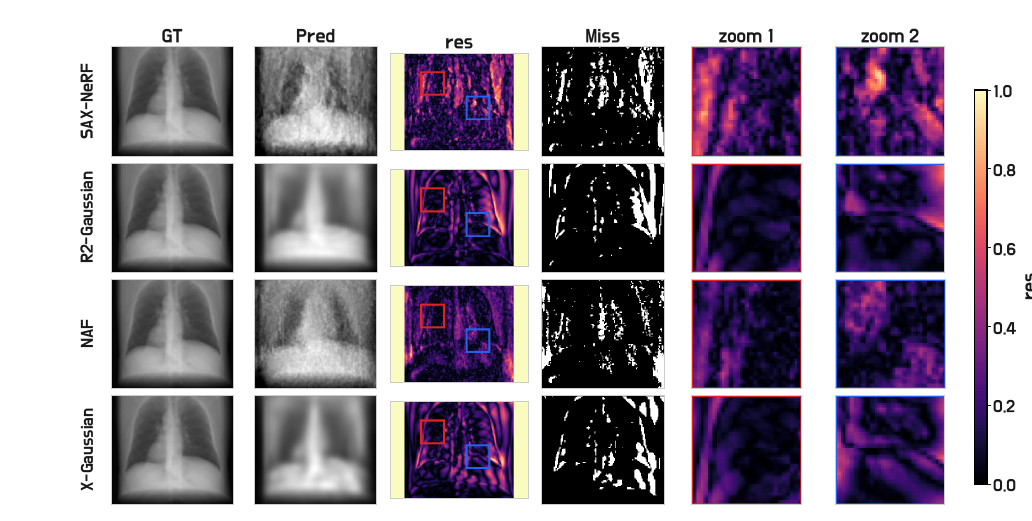}
\caption{External methods with residual zoom insets on chest far queries.
Colored boxes mark local regions.
The bottom-right zooms enlarge residual structure that a full-frame thumbnail hides.}
\label{fig:supp-external-zoom}
\end{figure*}

SAX-NeRF, R$^2$-Gaussian, NAF, and X-Gaussian stay sharp, and one calibration scale still puts most far pixels outside the interval.
\Cref{fig:supp-external-zoom} localizes that residual on chest edges and soft-tissue boundaries.
A constant scale cannot widen only at those boundaries.
The ACUS scale is spatially varying, and chest far coverage stays at $0.80$ to $0.87$.
These crops are not a per-method leaderboard and do not assign a scalar ACUS to the learned reconstructors.

\begin{figure*}[t]
\centering
\includegraphics[width=\textwidth]{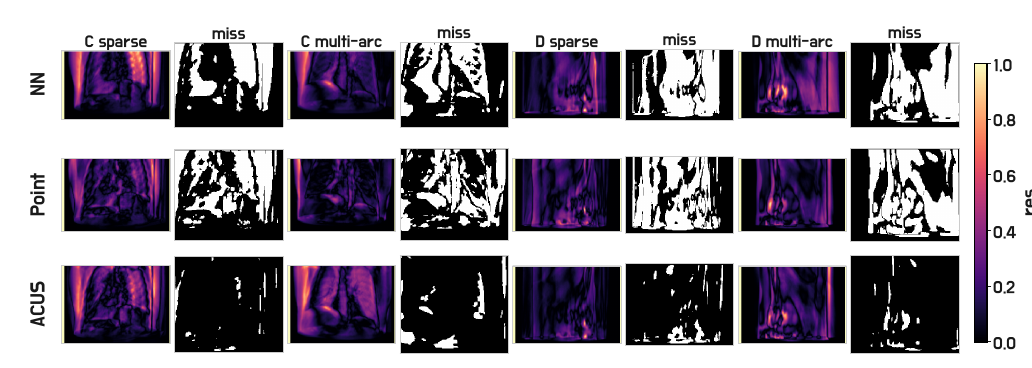}
\caption{Residual and miss mosaic across nearest neighbor (NN), point completion, and ACUS under four protocols.
Warm residual mass on far queries is the qualitative signature of an under-wide interval.}
\label{fig:supp-mosaic}
\end{figure*}

Rows of \cref{fig:supp-mosaic} are nearest neighbor, point completion, and ACUS; columns are the four anatomy-by-packing protocols.
On dental multi-arc far queries, point completion reaches $22.47\,\mathrm{dB}$ and ACUS $22.37\,\mathrm{dB}$, at scales $0.018$ and $0.063$.
The tenth of a decibel favors the constant-scale mean.
The miss mask does not: far coverage is reported as zero for point completion and $0.91$ for ACUS.
The same split appears in all four columns.

\section{Zoom-inset medical panels}
\label{sec:supp-zoom}

\begin{figure*}[t]
\centering
\includegraphics[width=\textwidth]{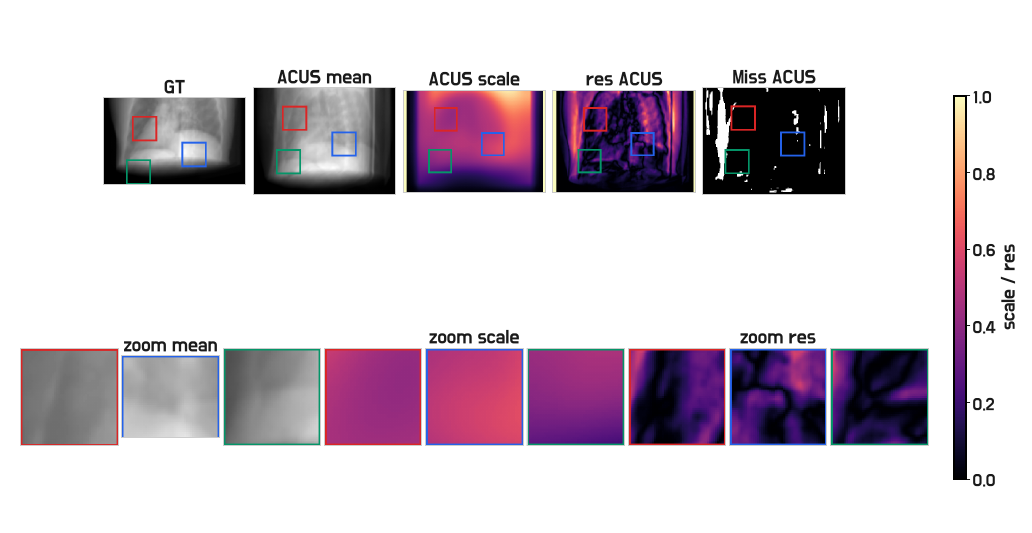}
\caption{Chest sparse overview with local zoom insets.
Top row: full panels with colored region-of-interest (ROI) boxes.
Bottom row: magnified crops of ACUS mean, scale, and residual.}
\label{fig:supp-zoom-chest-sparse}
\end{figure*}

\begin{figure*}[t]
\centering
\includegraphics[width=\textwidth]{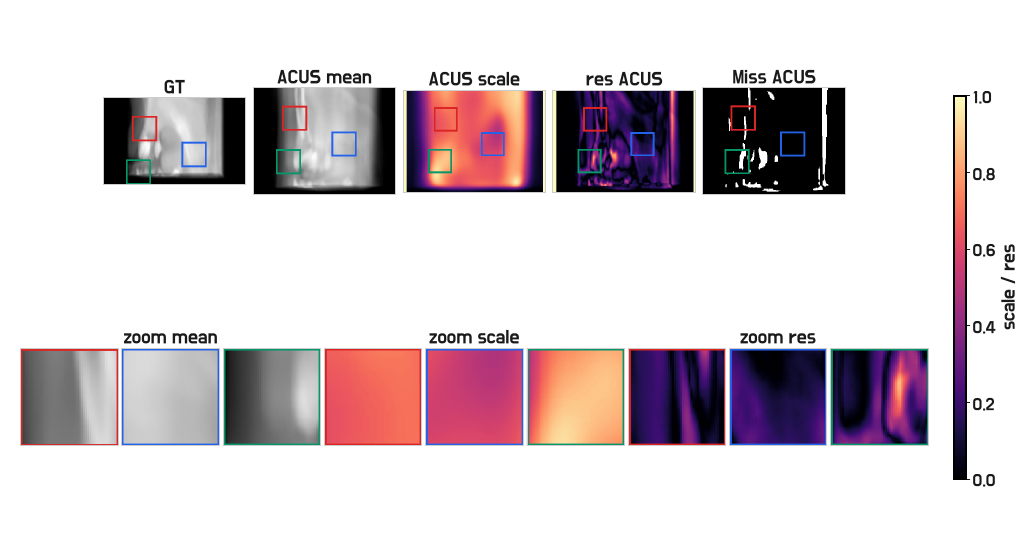}
\caption{Dental multi-arc overview with local zoom insets.
The dental scale is smaller in absolute value, and the residual zooms remain structured and locally organized.}
\label{fig:supp-zoom-dental}
\end{figure*}

The crops keep the marked regions and the display scale of the full panels.
They do not change a detector-wide rate.
Chest sparse far coverage is $0.87$ at PSNR $18.64\,\mathrm{dB}$, $13.8\,\mathrm{dB}$ below the near PSNR $32.48\,\mathrm{dB}$, which is why a far residual crop still shows anatomy-shaped error.
Chest multi-arc far coverage is $0.80$ at scale $0.095$.
Dental multi-arc far coverage is $0.91$ at scale $0.063$ and PSNR $22.37\,\mathrm{dB}$.
The smaller dental scale and the higher far PSNR are the same dental operating point as \cref{tab:supp-dental-arc-bins}.
The boxes were chosen to expose residual texture, not to sample the detector.
A clean crop can sit inside the covered majority, and a bright crop can be one of the misses whose outside distance, multiplied by $20$, would dominate the score.
Either crop is compatible with the main observation that correlated misses can concentrate on a small structure.

\begin{figure*}[t]
\centering
\includegraphics[width=\textwidth]{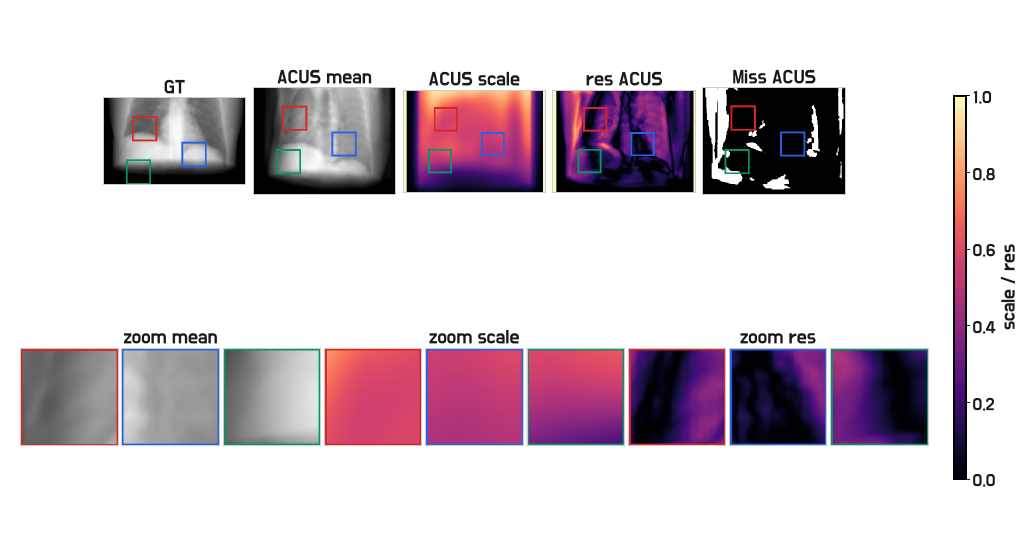}
\caption{Chest multi-arc overview with local zoom insets.
Comparing \cref{fig:supp-zoom-chest-sparse,fig:supp-zoom-dental,fig:supp-zoom-chest-multi} shows that the zoom pattern tracks the gap, not the packing of the acquired arcs.}
\label{fig:supp-zoom-chest-multi}
\end{figure*}

\section{Protocol walls and qualitative boards}
\label{sec:supp-walls}

\begin{figure*}[t]
\centering
\includegraphics[width=\textwidth]{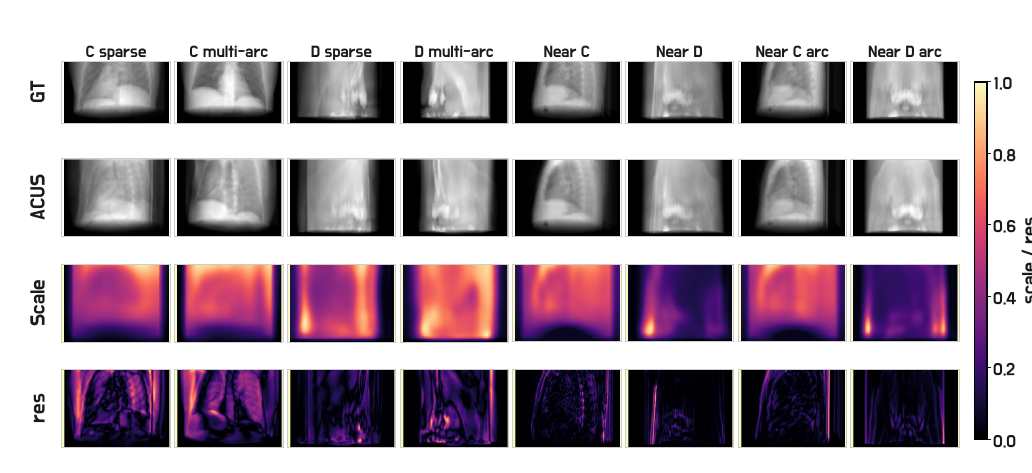}
\caption{Protocol wall across eight settings.
Rows: ground truth, ACUS mean, ACUS scale, absolute residual.
Columns: sparse, multi-arc, near, and mixed near protocols on chest and dental.}
\label{fig:supp-wall}
\end{figure*}

\Cref{fig:supp-wall} lines up ground truth, $\mu$, scale, and absolute residual for sparse, multi-arc, and near settings on both anatomies.
Near columns match near PSNR of $29.6$ to $32.5\,\mathrm{dB}$.
Far columns match far PSNR of $17.5$ to $22.4\,\mathrm{dB}$.
The scale row stays dark at the headline near scales $0.018$ to $0.027$ and opens at the headline far scales $0.061$ to $0.095$, with the thin chest tail at $0.111$.
In domain, scale and residual open together.
A dark scale beside a bright residual is the constant-scale and dental-to-chest pattern, and it is absent from this wall.

\begin{figure*}[t]
\centering
\includegraphics[width=\textwidth]{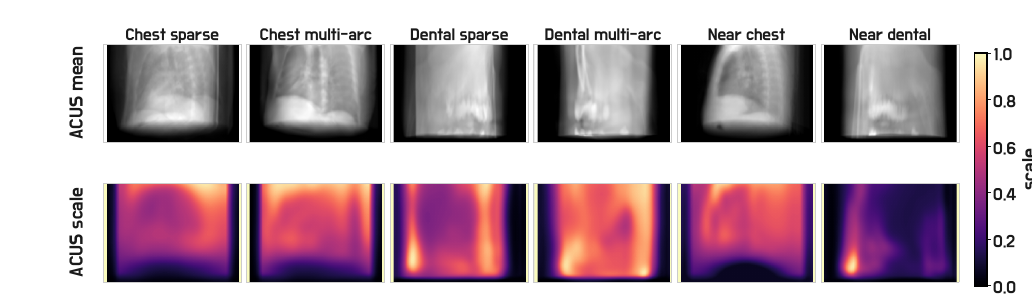}
\caption{ACUS scale heatmap wall across six protocols.
The scale opens on far settings and stays dark near acquired views.}
\label{fig:supp-heatmap}
\end{figure*}

\Cref{fig:supp-heatmap} removes the mean and the residual.
Dental stays cooler than chest, $0.061$ against $0.088$ sparse and $0.063$ against $0.095$ multi-arc.
Chest is still rising from the multi-arc $40^\circ$--$60^\circ$ bin ($0.094$) to the thin tail ($0.111$).
Dental has flattened, $0.062$ then $0.068$.
The widest mean, $0.111$, is about one third of the per-pixel ceiling $0.35$, so the opening is not an artifact of the clamp.

\begin{figure*}[t]
\centering
\includegraphics[width=0.88\textwidth]{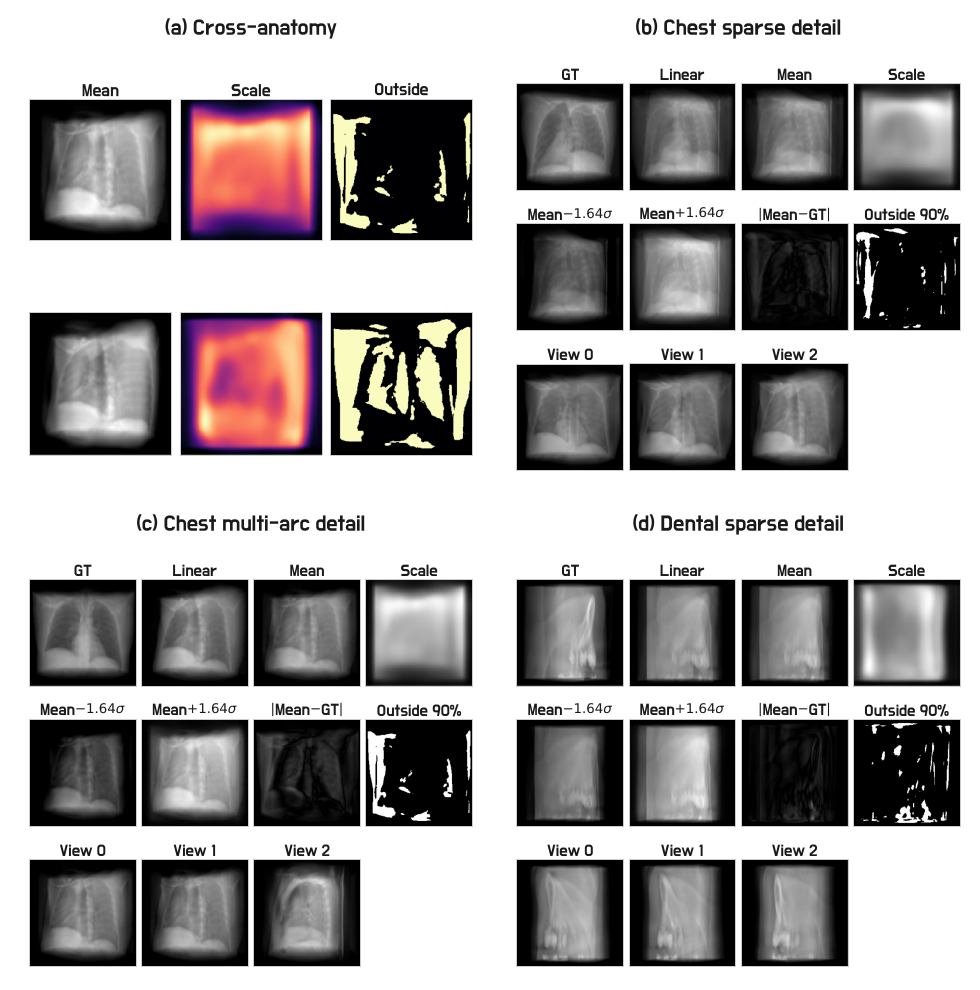}
\caption{Cross-anatomy and detailed qualitative panels.
(a) Cross-anatomy transfer.
(b) to (d) Domain-specific detail boards.}
\label{fig:supp-cross}
\end{figure*}

Panel (a) of \cref{fig:supp-cross} is \cref{tab:supp-transfer}.
Chest to dental keeps far width $0.093{\pm}0.013$, coverage $0.84$, and PSNR $19.59{\pm}2.08$.
That width sits with the chest far scales $0.088$ (sparse), $0.095$ (multi-arc), and $0.090$ (long arc), not with the dental far scale $0.063$.
Dental to chest keeps far width $0.061{\pm}0.001$, coverage $0.59$, and PSNR $16.58{\pm}1.12$.
The width matches the dental sparse far scale $0.061$.
Coverage falls to $0.59$, from in-domain chest far coverage of $0.80$ on the matched multi-arc protocol and $0.87$ on sparse orbits.
This is the overconfident cell in the main text: small scale, failed coverage.
The chest-to-dental width deviation $0.013$ is larger than the in-domain chest far deviations in the main captions ($0.003$ sparse, $0.008$ multi-arc).
The dental-to-chest deviation $0.001$ stays tight.
Panels (b) to (d) are matched-specialist boards and are not transfer results.

\begin{figure*}[t]
\centering
\includegraphics[width=\textwidth]{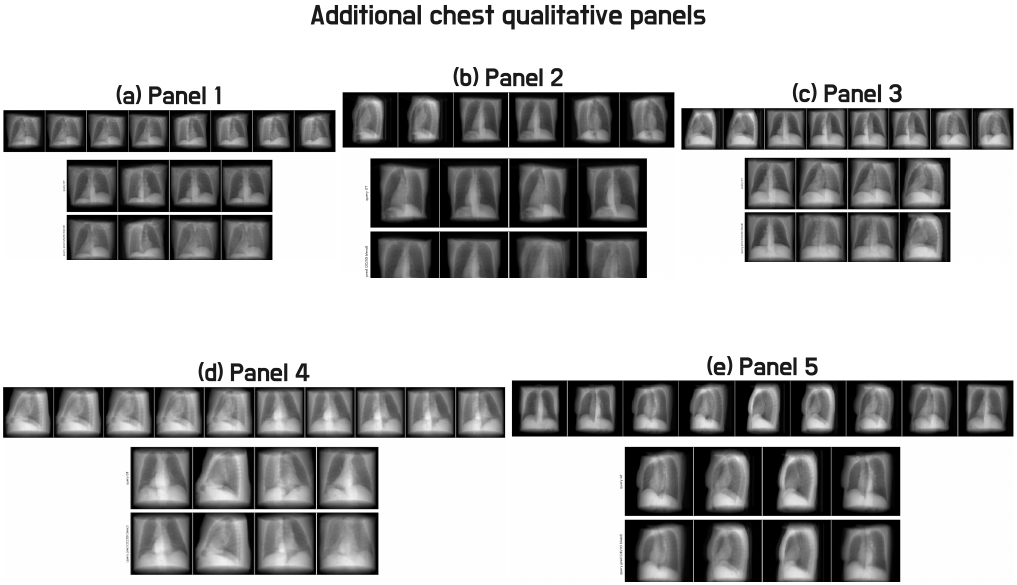}
\caption{Additional chest qualitative panels from held-out orbits.
(a) to (e) are distinct cases, each showing a query strip, the held-out ground-truth projection, and the predicted projection.
These panels are included to show breadth across cases, not to claim a new metric.}
\label{fig:supp-extra-chest}
\end{figure*}

\Cref{fig:supp-extra-chest} shows five further chest orbits.
The predicted projections keep the large structures and lose detail, consistent with far PSNR near $18\,\mathrm{dB}$ rather than near PSNR near $32\,\mathrm{dB}$.
Five cases do not change the 40-case rates in \cref{tab:supp-chest-sparse-bins,tab:supp-chest-arc-bins}.

\section{Orbit, volume, and collapse diagnostics}
\label{sec:supp-3d}

\begin{figure*}[t]
\centering
\includegraphics[width=\textwidth]{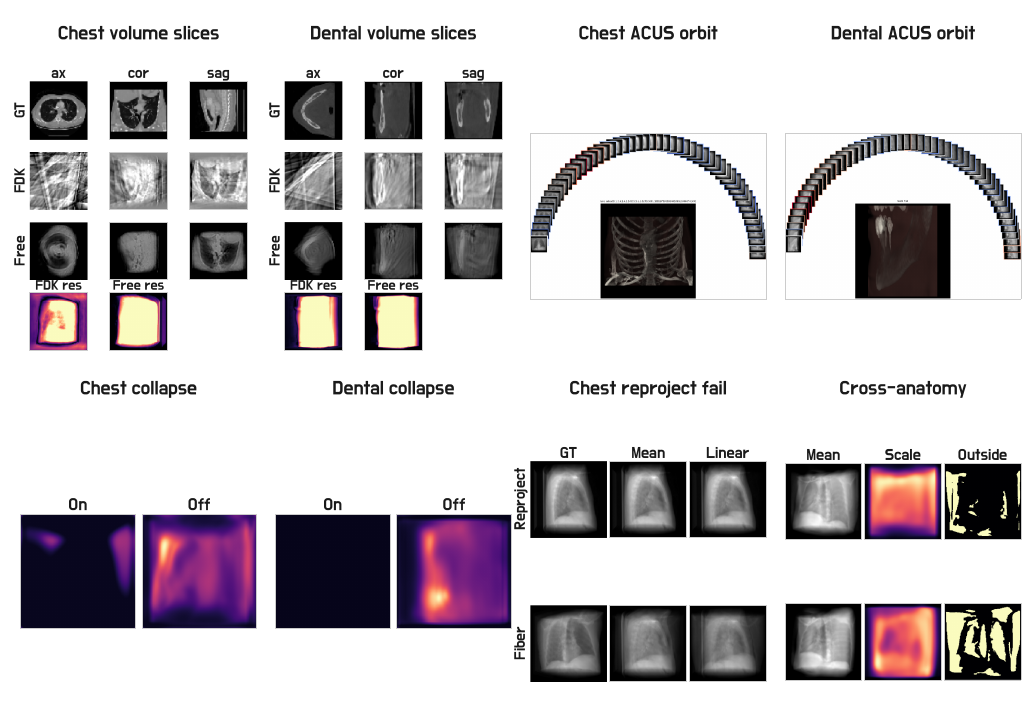}
\caption{Combined 3D orbit, volume-slice, collapse, and cross-anatomy board.
The volume slice row is a negative diagnostic from the main text.
Higher volume PSNR can coexist with worse agreement on acquired projections.}
\label{fig:supp-3d-board}
\end{figure*}

The orbit ring in main \cref{fig:orbit-rings} is the geometric reading of ACUS.
Each surrounding projection is bordered by its score.
Cold borders are near queries, at official near scales $0.018$ to $0.027$ and near PSNR $29.6$ to $32.5\,\mathrm{dB}$.
Warm borders are far queries, at far scales $0.061$ to $0.095$ and far PSNR $17.5$ to $22.4\,\mathrm{dB}$.
Acquired views themselves are narrower still, at scale $0.002$ (\cref{tab:supp-acq-width}).
The ring is \cref{fig:gap} drawn on the gantry.
The center is a spatial anchor, not a volume in the measurement fiber.
Main \cref{fig:orbit-schematics} marks which angles were kept.

\begin{figure*}[t]
\centering
\includegraphics[width=0.88\textwidth]{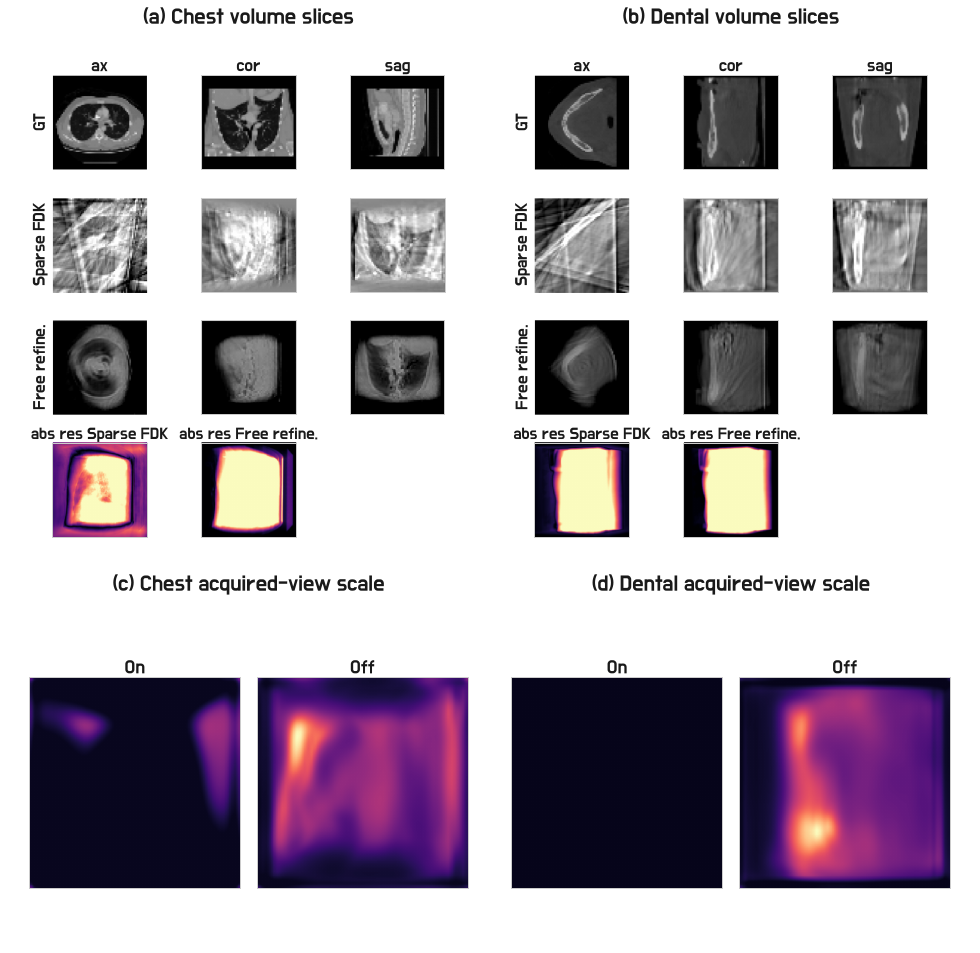}
\caption{Volume-slice diagnostic and acquired-view collapse maps.
(a) and (b) Axial and related slices from the $64^3$ free residual refinement.
(c) and (d) Acquired-view scale maps that should collapse near zero.}
\label{fig:supp-volume-collapse}
\end{figure*}

\begin{figure*}[t]
\centering
\includegraphics[width=\textwidth]{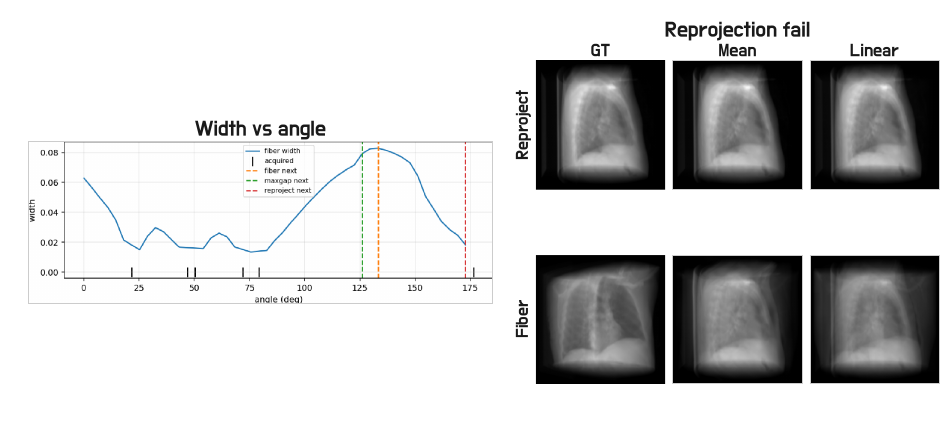}
\caption{Width-versus-angle curve and reprojection diagnostic for the volume readout experiment.}
\label{fig:supp-reproj}
\end{figure*}

On the 12 sparse cases behind main \cref{fig:volume-slices} and \cref{tab:supp-volume}, chest volume PSNR rises from $6.82{\pm}0.36$ to $14.04{\pm}1.11$ and reprojection PSNR falls from $13.32{\pm}1.41$ to $9.79{\pm}1.10$.
Dental volume PSNR rises from $9.41{\pm}0.56$ to $18.26{\pm}0.82$ and reprojection PSNR falls from $10.41{\pm}0.99$ to $9.13{\pm}0.87$.
The gains are $7.2\,\mathrm{dB}$ and $8.9\,\mathrm{dB}$.
The losses are $3.5\,\mathrm{dB}$ and $1.3\,\mathrm{dB}$.
The refinement was not trained with query-angle likelihood, and the FDK intensities are not on the interval's measurement scale, so the FDK inclusion cell is marked n/a.
Only $0.07$ to $0.15$ of the refined query-angle forward projections fall inside the predicted intervals, against far coverage of $0.80$ to $0.91$ when the same intervals are scored on held-out ground truth.
\Cref{fig:supp-volume-collapse} separates those slices from acquired-view scale maps, which belong with the acquired PSNR of $46.4$ to $54.9\,\mathrm{dB}$ in \cref{tab:supp-acq-width}.
\Cref{fig:supp-reproj} plots width against angle beside the reprojection curve that falls while volume PSNR rises.

\section{Coverage and acquired-view diagnostics}
\label{sec:supp-reliability}

\begin{figure*}[t]
\centering
\includegraphics[width=0.95\textwidth]{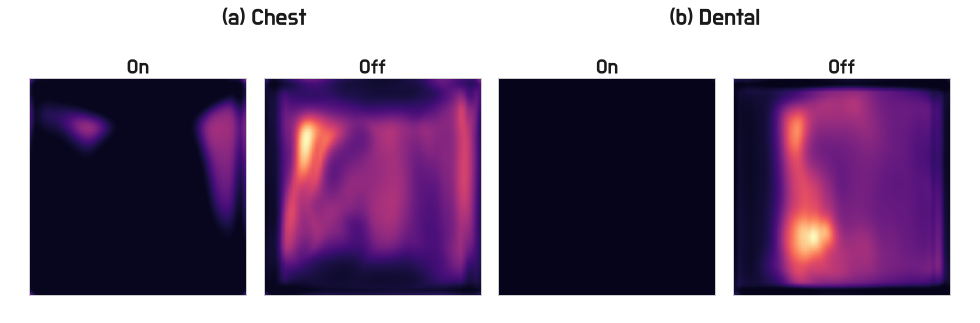}
\caption{Collapse gate on acquired views for chest and dental specialists.}
\label{fig:supp-collapse-maps}
\end{figure*}

\begin{figure*}[t]
\centering
\includegraphics[width=0.92\textwidth]{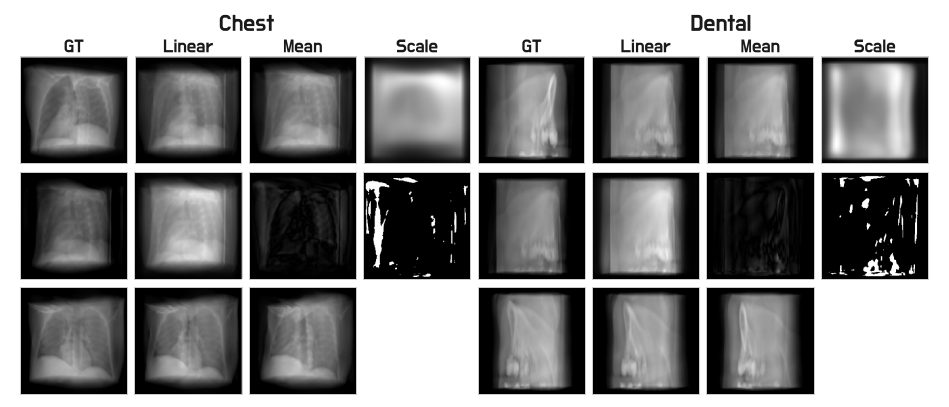}
\caption{Sparse far-query detail boards for chest and dental.}
\label{fig:supp-far-sparse}
\end{figure*}

The $77\%$ to $92\%$ range in the main conclusion is the eight headline cells: chest sparse $0.81/0.87$, chest multi-arc $0.77/0.80$, dental sparse $0.92/0.86$, dental multi-arc $0.89/0.91$.
Bins extend past it, from chest multi-arc $0^\circ$--$10^\circ$ at $0.750{\pm}0.062$ to dental sparse $0^\circ$--$10^\circ$ at $0.931{\pm}0.024$.
These are untempered Gaussian frequencies, not a distribution-free guarantee.
The widest sparse-chest coverage spread is ${\pm}0.087$ on the $30^\circ$--$40^\circ$ bin.
The $60^\circ$--$180^\circ$ row combines the largest chest scale, $0.111$, with a displayed deviation of zero because the row is thin.

\begin{figure*}[t]
\centering
\includegraphics[width=\textwidth]{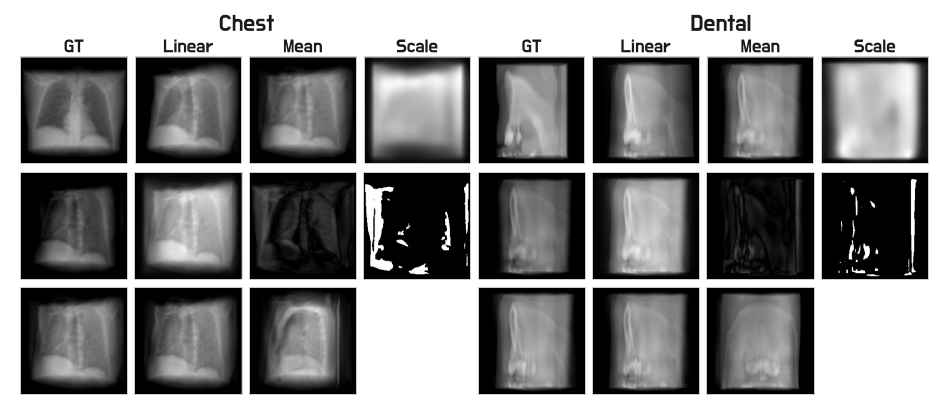}
\caption{Multi-arc far-query detail boards for chest and dental.}
\label{fig:supp-far-multi}
\end{figure*}

\Cref{fig:supp-collapse-maps} is the acquired-view gate in \cref{tab:supp-acq-width}.
Predictive PSNR is $46.4$, $52.6$, $49.5$, and $54.9\,\mathrm{dB}$ on chest sparse, chest multi-arc, dental sparse, and dental multi-arc, which is $27.8$ to $35.1\,\mathrm{dB}$ above the far PSNR of the same cell.
Removing $\mathcal{L}_{\mathrm{col}}$ on the 20-case split multiplies acquired width by nine on chest ($0.002$ to $0.018$) and by $7.5$ on dental ($0.002$ to $0.015$).
Far width is unchanged, and far coverage moves only from $0.80$ to $0.79$ and from $0.90$ to $0.89$.
\Cref{fig:supp-far-sparse,fig:supp-far-multi} add far-query boards under both packings.
They use the same miss convention as main \cref{fig:medfar} and do not change the 40-case estimates.

\section{Extended tables}
\label{sec:supp-tables}

The four bin tables reprint main \cref{tab:bins} in a one-domain-per-table layout so that a curve, a mid-gap figure, and a headline cell can be checked without reconstructing the pool by hand.
The remaining tables are the numerical form of the side experiments: analytical reprojection, collapse ablation, cross-anatomy transfer, the long-arc stress test, the volume readout, and the acquired-view gate.
This section states how each table lines up with a sentence in the main results or the main discussion.
The prose in \cref{sec:supp-curves,sec:supp-3d,sec:supp-reliability} already walks the bin, volume, and acquired-view numbers.
The paragraphs after the tables take the side experiments one at a time.

\begin{table}[t]
\centering
\caption{Chest sparse bins (40 cases), from the same official $K{=}8$ split as the main sparse table. $u$ is mean scale, not interval length. The $40^\circ$ to $60^\circ$ row is the far column of that table.}
\label{tab:supp-chest-sparse-bins}
\scriptsize
\begin{tabular*}{\columnwidth}{@{\extracolsep{\fill}}lccc@{}}
\toprule
Gap & $u$ & Cover & PSNR \\
\midrule
$0^\circ$ to $10^\circ$ & $0.014{\pm}0.001$ & $0.791{\pm}0.069$ & $33.88{\pm}1.56$ \\
$10^\circ$ to $20^\circ$ & $0.027{\pm}0.005$ & $0.848{\pm}0.063$ & $29.49{\pm}2.47$ \\
$20^\circ$ to $30^\circ$ & $0.055{\pm}0.009$ & $0.851{\pm}0.066$ & $23.11{\pm}1.71$ \\
$30^\circ$ to $40^\circ$ & $0.079{\pm}0.004$ & $0.833{\pm}0.087$ & $19.18{\pm}1.50$ \\
$40^\circ$ to $60^\circ$ & $0.088{\pm}0.003$ & $0.869{\pm}0.052$ & $18.64{\pm}0.99$ \\
\bottomrule
\end{tabular*}
\end{table}

\begin{table}[t]
\centering
\caption{Chest multi-arc bins (40 cases), from the same official split as the main multi-arc table. $u$ is mean scale, not interval length. Far queries aggregate the $40^\circ$ to $60^\circ$ and $60^\circ$ to $180^\circ$ bins.}
\label{tab:supp-chest-arc-bins}
\scriptsize
\begin{tabular*}{\columnwidth}{@{\extracolsep{\fill}}lccc@{}}
\toprule
Gap & $u$ & Cover & PSNR \\
\midrule
$0^\circ$ to $10^\circ$ & $0.013{\pm}0.002$ & $0.750{\pm}0.062$ & $32.88{\pm}1.30$ \\
$10^\circ$ to $20^\circ$ & $0.034{\pm}0.005$ & $0.785{\pm}0.066$ & $25.75{\pm}1.65$ \\
$20^\circ$ to $30^\circ$ & $0.057{\pm}0.008$ & $0.859{\pm}0.053$ & $23.15{\pm}1.55$ \\
$30^\circ$ to $40^\circ$ & $0.075{\pm}0.012$ & $0.868{\pm}0.062$ & $21.10{\pm}2.31$ \\
$40^\circ$ to $60^\circ$ & $0.094{\pm}0.008$ & $0.800{\pm}0.054$ & $17.54{\pm}0.53$ \\
$60^\circ$ to $180^\circ$ & $0.111{\pm}0.000$ & $0.804{\pm}0.000$ & $16.68{\pm}0.00$ \\
\bottomrule
\end{tabular*}
\end{table}

\begin{table}[t]
\centering
\caption{Dental sparse bins (40 cases), from the same official split as the main sparse table. $u$ is mean scale, not interval length. The $40^\circ$ to $60^\circ$ row is the far column of that table.}
\label{tab:supp-dental-sparse-bins}
\scriptsize
\begin{tabular*}{\columnwidth}{@{\extracolsep{\fill}}lccc@{}}
\toprule
Gap & $u$ & Cover & PSNR \\
\midrule
$0^\circ$ to $10^\circ$ & $0.021{\pm}0.001$ & $0.931{\pm}0.024$ & $33.17{\pm}1.04$ \\
$10^\circ$ to $20^\circ$ & $0.030{\pm}0.002$ & $0.911{\pm}0.043$ & $29.15{\pm}1.83$ \\
$20^\circ$ to $30^\circ$ & $0.047{\pm}0.004$ & $0.900{\pm}0.047$ & $24.71{\pm}1.98$ \\
$30^\circ$ to $40^\circ$ & $0.057{\pm}0.001$ & $0.860{\pm}0.051$ & $22.00{\pm}0.57$ \\
$40^\circ$ to $60^\circ$ & $0.061{\pm}0.002$ & $0.858{\pm}0.083$ & $21.12{\pm}2.04$ \\
\bottomrule
\end{tabular*}
\end{table}

\begin{table}[t]
\centering
\caption{Dental multi-arc bins (40 cases), from the same official split as the main multi-arc table. $u$ is mean scale, not interval length. Far queries aggregate the $40^\circ$ to $60^\circ$ and $60^\circ$ to $180^\circ$ bins.}
\label{tab:supp-dental-arc-bins}
\scriptsize
\begin{tabular*}{\columnwidth}{@{\extracolsep{\fill}}lccc@{}}
\toprule
Gap & $u$ & Cover & PSNR \\
\midrule
$0^\circ$ to $10^\circ$ & $0.021{\pm}0.001$ & $0.914{\pm}0.029$ & $32.70{\pm}1.27$ \\
$10^\circ$ to $20^\circ$ & $0.034{\pm}0.002$ & $0.853{\pm}0.055$ & $26.38{\pm}1.40$ \\
$20^\circ$ to $30^\circ$ & $0.047{\pm}0.003$ & $0.862{\pm}0.064$ & $24.16{\pm}1.51$ \\
$30^\circ$ to $40^\circ$ & $0.055{\pm}0.003$ & $0.870{\pm}0.067$ & $22.85{\pm}1.68$ \\
$40^\circ$ to $60^\circ$ & $0.062{\pm}0.002$ & $0.907{\pm}0.025$ & $22.41{\pm}0.81$ \\
$60^\circ$ to $180^\circ$ & $0.068{\pm}0.000$ & $0.914{\pm}0.000$ & $21.59{\pm}0.00$ \\
\bottomrule
\end{tabular*}
\end{table}

\begin{table}[t]
\centering
\caption{Same-geometry analytical reconstructors on chest: Feldkamp-Davis-Kress (FDK)~\cite{feldkamp1984fdk}, Simultaneous Algebraic Reconstruction Technique (SART)~\cite{andersen1984sart}, and adaptive-steepest-descent projection onto convex sets (ASD-POCS)~\cite{sidky2008asdpocs}.}
\label{tab:supp-analytical}
\scriptsize
\begin{tabular*}{\columnwidth}{@{\extracolsep{\fill}}lcccc@{}}
\toprule
Method & Scale & Cover near/far & PSNR near/far & ACUS \\
\midrule
FDK & $0.144$ & $0.57 / 0.50$ & $11.13 / 11.60$ & $0.739$ \\
SART & $0.086$ & $0.60 / 0.50$ & $15.58 / 14.51$ & $0.509$ \\
ASD-POCS & $0.087$ & $0.60 / 0.50$ & $15.57 / 14.51$ & $0.509$ \\
\bottomrule
\end{tabular*}
\end{table}

\begin{table}[t]
\centering
\caption{Collapse ablation on a 20-case multi-arc split.}
\label{tab:supp-collapse}
\scriptsize
\begin{tabular*}{\columnwidth}{@{\extracolsep{\fill}}lcccc@{}}
\toprule
Domain & Setting & Far width & Acq.\ width & Far cover \\
\midrule
Chest & with $\mathcal{L}_{\mathrm{col}}$ & $0.095$ & $0.002$ & $0.80$ \\
Chest & without $\mathcal{L}_{\mathrm{col}}$ & $0.095$ & $0.018$ & $0.79$ \\
Dental & with $\mathcal{L}_{\mathrm{col}}$ & $0.071$ & $0.002$ & $0.90$ \\
Dental & without $\mathcal{L}_{\mathrm{col}}$ & $0.071$ & $0.015$ & $0.89$ \\
\bottomrule
\end{tabular*}
\end{table}

\begin{table}[t]
\centering
\caption{Cross-anatomy transfer on multi-arc orbits.}
\label{tab:supp-transfer}
\scriptsize
\begin{tabular*}{\columnwidth}{@{\extracolsep{\fill}}lccc@{}}
\toprule
Train $\to$ Test & Far width & Far cover & Far PSNR \\
\midrule
Chest $\to$ Dental & $0.093{\pm}0.013$ & $0.84$ & $19.59{\pm}2.08$ \\
Dental $\to$ Chest & $0.061{\pm}0.001$ & $0.59$ & $16.58{\pm}1.12$ \\
\bottomrule
\end{tabular*}
\end{table}

\begin{table}[t]
\centering
\caption{Long-arc stress test (16 cases, one arc of $50^\circ$ to $120^\circ$).}
\label{tab:supp-longarc}
\scriptsize
\begin{tabular*}{\columnwidth}{@{\extracolsep{\fill}}lcccc@{}}
\toprule
Domain & Near width & Far width & PSNR near/far & Far cover \\
\midrule
Chest & $0.027$ & $0.090$ & $32.45 / 18.68$ & $0.82$ \\
Dental & $0.027$ & $0.064$ & $31.25 / 21.28$ & $0.88$ \\
\bottomrule
\end{tabular*}
\end{table}

\begin{table}[t]
\centering
\caption{Volume readout diagnostic on 12 sparse cases at $64^3$.}
\label{tab:supp-volume}
\scriptsize
\begin{tabular*}{\columnwidth}{@{\extracolsep{\fill}}lcccc@{}}
\toprule
Domain & Method & Vol.\ PSNR & Reproj.\ PSNR & Query interval inclusion \\
\midrule
Chest & sparse FDK & $6.82{\pm}0.36$ & $13.32{\pm}1.41$ & n/a \\
Chest & free residual refinement & $14.04{\pm}1.11$ & $9.79{\pm}1.10$ & $0.07$ to $0.15$ \\
Dental & sparse FDK & $9.41{\pm}0.56$ & $10.41{\pm}0.99$ & n/a \\
Dental & free residual refinement & $18.26{\pm}0.82$ & $9.13{\pm}0.87$ & $0.07$ to $0.15$ \\
\bottomrule
\end{tabular*}
\end{table}

\begin{table}[t]
\centering
\caption{Acquired-view collapse check for ACUS (main $K{=}8$ cells).}
\label{tab:supp-acq-width}
\scriptsize
\begin{tabular*}{\columnwidth}{@{\extracolsep{\fill}}llcc@{}}
\toprule
Domain & Orbit & Acq.\ width & Pred.\ PSNR on acquired \\
\midrule
Chest & Sparse & $0.002$ & $46.4$ \\
Chest & Multi-arc & $0.002$ & $52.6$ \\
Dental & Sparse & $0.002$ & $49.5$ \\
Dental & Multi-arc & $0.002$ & $54.9$ \\
\bottomrule
\end{tabular*}
\end{table}

\Cref{tab:supp-analytical} is read in \cref{sec:supp-external}.
\Cref{tab:supp-collapse,tab:supp-transfer} are read with \cref{fig:supp-winkler,fig:supp-cross}.
The long-arc table is the remaining comparison.
On 16 cases and one arc of $50^\circ$ to $120^\circ$, chest near/far width is $0.027/0.090$, near/far PSNR is $32.45/18.68$, and far coverage is $0.82$.
Dental near/far width is $0.027/0.064$, near/far PSNR is $31.25/21.28$, and far coverage is $0.88$.
The far scales sit with the official far scales of the same anatomy ($0.088$ and $0.095$ on chest, $0.061$ and $0.063$ on dental).
Sixteen cases are not pooled into main \cref{tab:sparse,tab:arc}.
Copy baselines are absent from \cref{tab:supp-acq-width} because an acquired-angle query returns the measurement exactly.

\section{Gap-binned constant scales}
\label{sec:supp-binned}

The main tables assign nearest neighbor, linear interpolation, and point completion one residual scale, fit on the eight calibration cases and then frozen at every query.
That design matches the failure drawn in main \cref{fig:idea}(c): a static scale can cover near angles and miss far ones.
A natural objection is that the zero far coverage is an artifact of refusing the baseline a coarser calibration.
If each method may use a different positive constant in each gap bin, the miss penalty might disappear while the point predictor is left unchanged.

\begin{figure}[t]
\centering
\includegraphics[width=\columnwidth]{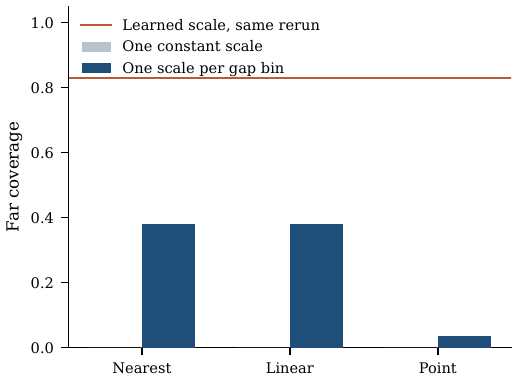}
\caption{Chest multi-arc far coverage on the later rerun, $N{=}40$, $K{=}8$.
Gray bars use one residual scale for every angle.
Blue bars use one residual scale per gap bin.
The line is the learned scale from that same rerun, not the official multi-arc cell.}
\label{fig:supp-binned}
\end{figure}

\Cref{fig:supp-binned} tests that objection on a later chest multi-arc rerun with the same view budget, $N{=}40$ and $K{=}8$.
One scale per bin raises nearest-neighbor and linear far coverage from $0$ to $0.381$.
Point completion, whose far PSNR is the highest of the three, reaches only $0.036$.
The learned scale on that same rerun keeps far coverage $0.829$.
Bin-wise constants therefore recover part of the coverage that a single constant loses, and they do not reproduce the learned coverage.
The mechanism is the same one in the main score.
A constant, even a different constant in each bin, has no pixelwise structure, so the pixels that sit far outside the band still dominate the miss term.
The learned scale opens with the gap and varies across the detector, which is what keeps coverage near the nominal rate while the point-estimate PSNR continues to fall.

On the sparse split of the same rerun the objection fails more sharply.
All three binned constants still have far coverage $0$.
The calibration residuals in those far bins do not open the scale above the near-angle magnitude, so a gap-indexed constant collapses back to the static scale the main tables already reject.

This rerun is not the official 40-case cell and is not pooled with \cref{tab:supp-chest-sparse-bins,tab:supp-chest-arc-bins}.
Its learned chest sparse far scale is $0.066$ with coverage $0.831$, against the official $0.088$ and $0.869$.
Its learned multi-arc far scale is $0.096$ with coverage $0.829$, against the official $0.095$ and $0.80$.
The comparison is internal to the rerun.
It supports the main claim that far coverage is a property of a gap-dependent scale, not a property that any positive constant can be tuned to match, and it does not replace the official numbers.

\section{Implementation and reproducibility notes}
\label{sec:supp-impl}

\paragraph{Architecture.}
The encoder is a frozen MAE ViT-B/16~\cite{he2022mae} on $224\times224$ projections, with learning rate zero.
Fusion uses the two nearest acquired views at softmax temperature $0.10$.
$\delta(a')$ still uses the full mask.
Matched bins of sparse and multi-arc orbits nevertheless have similar scales, which is the empirical content of that split: the unused views change $\mathcal{F}$ and do not enter the features.
The scale head, the only module added over point completion, is initialized near $\log 0.08$, between the official near scales near $0.02$ and the official far scales $0.06$ to $0.10$.
It is clamped to $\sigma\in[e^{-6},0.35]$.
The widest reported mean scale is $0.111$, about one third of the ceiling.
Near-angle mixing is off, so a narrow near interval comes from the scale head and $\mathcal{L}_{\mathrm{col}}$, not from copying a neighbor into $\mu$.

\paragraph{Optimization and splits.}
Each specialist trains for $8{,}000$ steps at batch size $1$ from a point-completion initialization.
$\mathcal{L}_{\mathrm{NLL}}$ and $\mathcal{L}_{\mathrm{col}}$ have equal weights.
\Cref{tab:supp-collapse} is the check on that choice: removing $\mathcal{L}_{\mathrm{col}}$ widens the acquired scale and leaves the far scale in place.
The $\ell_1$ and gradient penalties on $\mu$ use weights $0.15$ and $0.05$.
The scale has no ground-truth variance target.
Training draws $K$ from 6 to 10, so the test budget $K{=}8$ sits inside the training range, uses a sparse subset with probability $0.55$, holds out four queries, and drops $15\%$ of acquired views.
Chest and dental specialists are separate.
There is no shared head that would be forced to average a far scale near $0.06$ with one near $0.09$.
The eight calibration cases are disjoint from the 40 test cases.
The 20-case ablation, the 16-case long arc, and the 12-case volume readout are separate from that test set in role, including where a chest far number coincides.
A draw from the diagonal Gaussian need not be a projection of a volume in $\mathcal{F}$.
\Cref{sec:supp-3d} is the volume that fails both acquired-view agreement and query-angle inclusion.

\section{Reading guide for the figure set}
\label{sec:supp-guide}

\Cref{sec:supp-curves} reads the pooled cells against the bins, including the $32.4^\circ$ case and the 20-case collapse panel.
\Cref{sec:supp-nearfar,sec:supp-zoom} are the near/far and inset boards.
\Cref{sec:supp-external} is the reconstructor comparison and the only width/miss split of a printed scalar.
\Cref{sec:supp-walls} is the protocol wall, the scale heatmaps, and the transfer panels.
\Cref{sec:supp-3d,sec:supp-reliability} are the volume diagnostic, the acquired-view PSNR, and the extra far boards.
\Cref{sec:supp-tables} holds the numeric expansions, including the 16-case long arc.
\Cref{sec:supp-binned} is a later chest rerun and is not pooled with those tables.
\Cref{sec:supp-theory} is the scoring argument behind those tables.

\section{Interval scoring and consistency}
\label{sec:supp-theory}

\paragraph{Why the interval score targets two quantiles.}
For a scalar target $Y$ with a continuous cumulative distribution $F$ and finite first moment, write
\[
S_\alpha(L,U;Y)=(U-L)+\frac{2}{\alpha}\big[(L-Y)_++(Y-U)_+\big].
\]
The derivatives of the expected positive parts are $F(L)$ and $-[1-F(U)]$.
Consequently, stationarity gives $F(L)=\alpha/2$ and $F(U)=1-\alpha/2$.
The expected score is convex in each endpoint, so these quantiles minimize it. Unique quantiles give a unique optimum~\cite{winkler1972,gneiting2007}.
With atoms, the corresponding subgradient conditions permit non-unique quantiles.
Properness concerns the target quantiles. It is not strict propriety for the entire density, because distinct distributions can share both endpoints.
The symmetric Gaussian family restricts which pairs can be represented.
Properness of the score is therefore not evidence that a fitted Gaussian interval is calibrated.
The coverage columns are that evidence: $0.75$ to $0.93$ across bins, and $0.77$ to $0.92$ on the eight headline cells.
Chest remains below the nominal rate, which is why the main discussion asks for further calibration before $90\%$ is treated as an operating guarantee.

\paragraph{Sharpness and missed observations.}
At $\alpha=0.1$, the standard-normal quantile is $z_{0.95}\approx1.64485$.
The implemented rounded value $1.645$ gives interval length $3.29\sigma$ and an outside-distance multiplier of $20$.
Averaging is linear, so the detector-average score is exactly the average interval length plus $20$ times the average outside distance.
Coverage records only whether the outside distance is zero. It cannot identify the score without the magnitudes of the positive distances.
This is why the main tables are diagnostics of interval behavior rather than complete numerical ACUS rankings.
The factor of $20$ makes the missing magnitude consequential.
On the chest sparse far cell, $u{=}0.088$ and the width term is $3.29\times 0.088\approx 0.29$.
An average outside distance of about $0.015$, in the same intensity units as $\sigma$, produces a miss term of $20\times 0.015=0.30$ and already matches that entire width term.
On the dental sparse far cell, $u{=}0.061$ and the width term is about $0.20$, matched by an average outside distance of about $0.010$.
A constant-scale method with zero reported far coverage could therefore sit near the learned ACUS or several times above it, depending on how far outside the narrow band the pixels lie.
The miss masks show that those pixels exist and do not show the distance.
For the analytical rows the scalar is printed, and \cref{sec:supp-external} separates their width and miss terms.

\paragraph{A restricted continuity argument.}
Suppose a nonempty class $\mathcal B\cap\mathcal F$ satisfies a uniform angular modulus $\omega_{\mathcal B}$ in a fixed detector norm.
For two compatible volumes $V_1,V_2$ and a closest acquired angle $a$, both project to $y_a$ at $a$.
The triangle inequality gives
\[
\|P_{a'}(V_1)-P_{a'}(V_2)\|\leq 2\omega_{\mathcal B}(|a'-a|).
\]
Thus the restricted projection set contracts as the gap vanishes if the assumed modulus vanishes.
No modulus $\omega_{\mathcal B}$ is estimated.
The argument supplies the boundary condition tested by \cref{tab:supp-acq-width}, acquired scale $0.002$ and acquired PSNR $46.4$ to $54.9\,\mathrm{dB}$, and \cref{tab:supp-collapse} shows that $\mathcal{L}_{\mathrm{col}}$ enforces it.
The far curves are not that modulus.
Chest sparse opens from $0.014$ to $0.088$ and dental sparse from $0.021$ to $0.061$ on the same geometry, and \cref{tab:supp-transfer} keeps the geometry fixed while swapping the specialist.

\paragraph{What the training scale estimates.}
For a fixed conditioning input and mean, the expected Gaussian NLL has scale derivative $1/\sigma-\mathbb E[(Y-\mu)^2]/\sigma^3$.
Its unconstrained stationary scale is the conditional residual root-mean-square.
This includes error in the center and does not separate aleatoric uncertainty, epistemic uncertainty, and approximation bias.
The fitted network further imposes a scale floor and ceiling, parameter sharing, finite training, and auxiliary penalties.
Empirical coverage must therefore be evaluated rather than inferred from the likelihood objective.
The collapse loss tests a different property at acquired angles and cannot calibrate missing angles by itself.
The headline cells show what a scale that does track residual size looks like, and where it does not.
PSNR is a statement about root-mean-square error of $\mu$.
For a fixed peak, a drop of $\Delta$ decibels multiplies that error by $10^{\Delta/20}$.
Chest sparse falls from $32.48\,\mathrm{dB}$ to $18.64\,\mathrm{dB}$, a factor of about $4.9$ in amplitude, and the mean scale rises from $0.018$ to $0.088$, also a factor of about $4.9$.
Chest multi-arc falls from $29.59\,\mathrm{dB}$ to $17.48\,\mathrm{dB}$, a factor of about $4.0$, and the scale rises from $0.023$ to $0.095$, a factor of about $4.1$.
Dental multi-arc falls from $29.78\,\mathrm{dB}$ to $22.37\,\mathrm{dB}$, a factor of about $2.3$, and the scale rises from $0.027$ to $0.063$, a factor of about $2.3$.
Dental sparse is the exception: PSNR falls from $31.90\,\mathrm{dB}$ to $21.12\,\mathrm{dB}$, a factor of about $3.5$, while the scale rises from $0.024$ to $0.061$, a factor of about $2.5$.
Three of the four headline cells are what the NLL stationary condition leads one to expect when the pixelwise scale follows residual magnitude.
The fourth is the gap between that stationary condition and a fitted, clamped, penalized mean of pixelwise scales.
None of the four is an identity, and none of them recovers coverage from PSNR.
A constant scale fitted where residuals are small, which is what the eight calibration cases supply to the copy baselines, is the same stationary condition applied to the wrong gap.
On chest sparse the constant sits near $0.012$ to $0.014$, the size of the nearest bin, and the far residual is several times larger.
Zero reported far coverage is the result the miss penalty was introduced to charge.

\end{document}